\pdfoutput=1
\documentclass[mnsc, nonblindrev]{Management-Science-template/informs3Arxiv} 

\OneAndAHalfSpacedXI 

\usepackage{subcaption}
\usepackage{graphicx}
\usepackage{booktabs}
\usepackage{amssymb}
\usepackage{bm}
\usepackage{bbm}
\usepackage{enumitem}
\setlist[itemize]{align=parleft,left=0pt..1em}
\usepackage{arydshln}
\usepackage[ruled,linesnumbered]{algorithm2e}
\usepackage{multirow}

\usepackage[labelfont=bf]{caption}
\usepackage{natbib}
 \bibpunct[, ]{(}{)}{,}{a}{}{,}%
 \def\bibfont{\small}%
 \def\newblock{\ }%

\newcommand{\ind}{\perp\!\!\!\!\perp} 

\TheoremsNumberedThrough     

\EquationsNumberedThrough    

\MANUSCRIPTNO{MS-INS-22-01576.R2}

\begin{document}


\RUNAUTHOR{De-Arteaga et al.}

\RUNTITLE{Leveraging Expert Consistency to Improve Algorithmic Decision Support}
\TITLE{Leveraging Expert Consistency to Improve Algorithmic Decision Support}

\ARTICLEAUTHORS{%
\AUTHOR{Maria De-Arteaga}
\AFF{University of Texas at Austin, Austin, TX\\\EMAIL{dearteaga@mccombs.utexas.edu}} 
\AUTHOR{Vincent Jeanselme}
\AFF{University of Cambridge, Cambridge, UK\\\EMAIL{vincent.jeanselme@mrc-bsu.cam.ac.uk}}
\AUTHOR{Artur Dubrawski, Alexandra Chouldechova}
\AFF{Carnegie Mellon University, Pittsburgh, PA}
} 

\ABSTRACT{Machine learning (ML) is increasingly being used to support high-stakes decisions. However, there is frequently a \emph{construct gap}: a gap between the construct of interest to the decision-making task and what is captured in proxies used as labels to train ML models. As a result, ML models may fail to capture important dimensions of decision criteria, hampering their utility for decision support. Thus, an essential step in the design of ML systems for decision support is selecting a target label among available proxies. In this work, we explore the use of historical expert decisions as a rich---yet also imperfect---source of information that can be combined with observed outcomes to narrow the construct gap. We argue that managers and system designers may be interested in learning from experts in instances where they exhibit consistency with each other, while learning from observed outcomes otherwise. We develop a methodology to enable this goal using information that is commonly available in organizational information systems. This involves two core steps. First, we propose an influence function-based methodology to estimate expert consistency indirectly when each case in the data is assessed by a single expert. Second, we introduce a label amalgamation approach that allows ML models to simultaneously learn from expert decisions and observed outcomes. Our empirical evaluation, using simulations in a clinical setting and real-world data from the child welfare domain, indicates that the proposed approach successfully narrows the construct gap, yielding better predictive performance than learning from either observed outcomes or expert decisions alone. }

\KEYWORDS{predictive algorithms, decision support, machine learning, design science, expert consistency}

\maketitle

\section{Introduction}
Across domains, experts routinely make predictions to inform the decisions that their job requires them to make. Recruiters assess the likelihood of a candidate's succeeding at a job if hired; judges assess the likelihood of recidivism when considering bail; and child abuse hotline call workers assess the risk of harm to a child when deciding whether a hotline call should prompt an investigation. 
Increasingly, machine learning (ML) is being used to aid experts in these predictions. 
This shift is largely motivated by research showing that ML and actuarial models can perform better than humans in making predictions ~\citep{meehl1954clinical,dawes1989clinical,grove2000clinical,kleinberg2017human, angelova2022algorithmic}. 
However, studies that show the superior predictive power of ML often make overly simplistic assumptions about the data used to train ML models. In particular, while expert decisions often involve navigating multidimensional, complex criteria~\citep{elster1992local,nosofsky2005procedural}, such work often implicitly assumes that there is an observed outcome that perfectly captures the decision objective and that it can thus be used as the target label that a model is trained to predict. 
In practice, however, there is frequently what we call a \emph{construct gap}: the construct of interest to the decision-making task is not measured, and only proxies are available. Typically, available proxies fall in one of two buckets: historical expert decisions and observed outcomes. Historical expert decisions refer to the decisions that experts previously made for the task at hand, without using an ML decision support tool. Using this as a target corresponds to attempting to ``automate" the experts. The downside of such an approach is that experts are prone to idiosyncrasies, biases and errors, so training an ML model to replicate prior decisions risks replicating these imperfections. For this reason, organizations often favor the use of observed outcomes, which refer to outcomes of relevance to the decision-making task, which typically occur downstream from the decision point. However, it is often the case that what is observed is only a proxy for a holistic decision-making criteria. 
 Thus, identifying a suitable target label among (imperfect) choices is a challenging and crucial step that managers and designers face when developing a decision support tool.  


We formalize the problem as follows. 
Let there be an \emph{unobserved} construct of interest $Y^c$ that decision makers try to predict. For each data instance, we assume that observational data contain a set of covariates $X$, an expert decision $D$, and an observed outcome $Y$, which constitutes an imperfect proxy for $Y^c$. 
For instance, in healthcare, a decision about which patients to refer to care management programs might most directly consider patients' evolving \emph{health care needs} ($Y^c$). However, widely used deployed models have been trained to predict patients'\emph{observed health costs} ($Y$)~\citep{obermeyer2019dissecting}. Similarly, call workers in child abuse hotlines---the focus domain in the empirical portion of this paper--- receive reports of child neglect or abuse and are tasked with deciding which cases should be screened in for further investigation by a social worker. The construct of interest is whether the call involves a \emph{child at risk of harm} ($Y^c$), while deployed ML decision support models~\citep{chouldechova2018case} are trained to predict the observed outcome ($Y$) of whether the  child experiences an \emph{out-of-home placement}---that is, whether the child is later removed from the home and placed into foster care. Additional examples from the literature of proxy outcomes, constructs of interest, and expert decisions are provided in Table~\ref{tbl:examples}.

In addition to recording the observed outcome, $Y$, historical records often include human decisions, $D$, indicating decision-makers' determinations in the absence of an ML decision support tool. 
In the previous examples, $D$ corresponds to 
a physician's decision to refer a patient to the care management program or to a call worker's decision to screen in a call for investigation. These decisions can be grounded in deep expertise and domain knowledge about the construct $Y^c$, but they can also be mistaken, noisy, or biased. As a result, both $Y$ and $D$ are imperfect proxies for $Y^c$. 
Finally, in some domains, a selective labels problem~\citep{lakkaraju2017selective} may also be present, meaning that the outcome $Y$ may be selectively observed, conditioned on $D$, resulting in a single proxy being available for some cases. For example, an out-of-home placement, $Y$, may only occur following a decision to screen in a call for investigation.

\begin{table}[ht]
\centering
\caption{Practical examples of construct gaps in scenarios where ML is being deployed for decision support. }
\label{tbl:examples}
\SingleSpacedXI
\footnotesize
\def\arraystretch{1.3}
\begin{tabular}{p{3.8cm} p{4.2cm} p{3.8cm} p{3.5cm}}
\toprule 
\textbf{Domain} & \textbf{Construct of interest ($Y^c$)} &\textbf{Observed outcome ($Y$)}  & \textbf{Human decision (D)}   \\ 
\midrule 
 \multirow{2}{*}{\shortstack[l]{Health management \\ \citep{obermeyer2019dissecting}}}  &
health needs & health spending & refer patient to care management program\\
 \multirow{2}{*}{\shortstack[l]{Hiring \\\citep{raghavan2020mitigating} }}  &
\multirow{2}{*}{\shortstack[l]{most qualified \\ candidate}} & \# sales / retention & hire candidate\\
\vspace{0.2cm}
\multirow{2}{*}{\shortstack[l]{Criminal justice \\ \citep{fogliato2021validity} }}& \vspace{0.2cm}
recidivism & \vspace{0.2cm} rearrest & \vspace{0.2cm}grant bail\\
\vspace{0.3cm}
\multirow{2}{*}{\shortstack[l]{Child abuse hotline \\\citep{chouldechova2018case} }}& \vspace{0.3cm}
risk to child & \vspace{0.3cm} \multirow{2}{*}{\shortstack[l]{out-of-home \\ placement }} & \vspace{0.3cm} screen in call for investigation\\
\bottomrule
\end{tabular}
\end{table}

Recent work has discussed risks of misalignment between $Y$ and $Y^c$, often referred to as a construct validity issue~\citep{jacobs2021measurement}, emphasizing the importance of appropriate problem formulation in the design of sociotechnical ML systems~\citep{passi2019problem,jacobs2021measurement,raghavan2020mitigating}. Researchers also have shown empirically that optimizing for observed outcomes that serve as proxies to decision criteria can have a number of negative consequences, including algorithmic bias~\citep{obermeyer2019dissecting} and undesirable drifts in decision-making criteria~\citep{green2021algorithmic}. Simultaneously, the risks of learning from $D$ have also been studied, with researchers highlighting that it can lead algorithms to reproduce and even amplify errors in human judgment~\citep{mullainathan2017does}. 
So far, however, the conversation primarily has centered on the importance of selecting an appropriate proxy for the construct of interest. What we lack are technical approaches that give practitioners tools to better approximate a construct of interest when no single variable constitutes a sufficiently good proxy.

\paragraph{Managerial contributions.} We propose that when only proxies of the construct of interest are available, managers and system designers may be interested in training algorithms that simultaneously learn from expert decisions and observed outcomes. Specifically, we suggest that when domain knowledge indicates that agreement among multiple experts can be seen as indicative of correctness, it may be desirable to learn from experts in instances where they exhibit consistency with each other, while learning from observed outcomes otherwise. The underlying intuition is that in many domains expert agreement---for instance, physicians agreeing on a diagnosis---can be seen as a sign that (1) experts are correct, and (2) current expert knowledge is sufficient to assess that case. Meanwhile, when experts disagree with each other, (1) a single expert decision is unlikely to be reliable as it may reflect the individual expert's idiosyncrasies and biases, and (2) current expertise may be insufficient to correctly assess the case, indicating an opportunity to learn from observed outcomes.    
We propose a methodology to simultaneously learn from human decisions and observed outcomes and empirically show its potential benefits.

\paragraph{Methodological contributions.}
We formalize the problem of learning from both observed proxy outcomes, $Y$, and human decisions, $D$, to better approximate the unobserved construct of interest, $Y^c$. 
We propose a methodology that relies on information that is commonly available in organizational information systems to enable learning from both $Y$ and $D$, under the assumption that decision consistency across experts for particular instances is indicative of correctness---that is, of alignment between $D$ and $Y^c$. Here, we define consistency as agreement across multiple experts regarding the decision for a given instance. The proposed methodology consists of two main steps: (1) estimating expert consistency, and (2) leveraging the estimated expert consistency to approximate $Y^c$ more closely than either $Y$ or $D$ alone can do. 

Step 1 involves two key challenges: (i) in many settings of interest, each instance is assessed by only a single decision maker; and (ii) the assignment of decision makers to cases can be non-random. These elements differ from settings, such as crowd sourcing and data labeling, where instances are commonly assessed by multiple, randomly assigned decision makers. For example, a given patient might only be seen by a single physician, and each child maltreatment hotline call is screened by a single call worker. In addition, which physician sees a patient is likely to depend on location and insurance coverage, and call workers who primarily screen calls in the evenings and on weekends may see a different case mix than those who are working during the school day. Our task is therefore to infer whether there is consistency \emph{across} experts---that is, whether they would \textit{hypothetically} agree on their decision for a given case---when we do not directly observe multiple decisions per case and when the decision maker that assesses each case may not be randomly assigned. 

Our proposed approach for inferring expert consistency trains a prediction model that aims to replicate historical expert decisions, $D$, from observed features, $X$. Then, the approach uses influence functions~\citep{cook1986assessment} to assess whether high-certainty predictions are driven by the historical decisions of multiple experts---and are thus indicative of consistency across multiple experts---or if they are driven by the decisions of one or a few experts. Our use of the influence functions constitute a measure of robustness, where robust decision predictions are ones being driven by historical decisions of multiple experts. Taken together, the two stages of the procedure provide us with an estimate of consistency of historical decisions by reflecting both the certainty of decisions---measured in terms of the estimated probability, $P(D \mid X)$---and the extent to which this certainty is reflective of multi-expert agreement.

After we have an estimate of expert consistency for each instance in the historical data using the procedure described, in Step 2 we use this information via label amalgamation to augment the observed labels $Y$ with the decisions $D$ for \textit{high-consistency cases}. In effect, this approach learns from experts' decisions when they display certainty and consistency and learns from observed outcomes otherwise. By learning from expert decisions when they are consistent, we are better able to assess risk correctly in cases that might warrant a particular decision because of factors captured in $D$ 
but not captured in the observed outcome, $Y$. 
Indeed, we can demonstrate theoretically that when expert consistency indicates correctness, the proposed label amalgamation approach yields a better estimate of $Y^c$, compared to the model that solely predict proxy label $Y$ or human decision $D$. 


\paragraph{Evaluation and empirical contributions.}
To evaluate the proposed methodology, we first validate its performance, robustness, and failure modes using semi-synthetic data. The data set we use is widely used and publicly available, ensuring these results can be reproduced. We compare the proposed approach against other methods that aim to learn from different sources of labels, learn from noisy labels, and one approach---learning to defer---that also aims to combine observed outcomes and human decisions, albeit with a different goal. The results show that the proposed method is the only one able to simultaneously outperform the alternatives of learning from either $Y$ or $D$. In particular, in the vast majority of scenarios, all other approaches yield worse results than learning from $Y$ alone, indicating that they are unable to leverage expert knowledge when experts are able to assess some types of instances with outstanding performance, while being routinely incorrect or biased for other instances. 
We then turn our focus to the child welfare domain, where predictive models are increasingly being deployed to assist call workers in identifying high-risk cases that should be screened in for investigation~\citep{chouldechova2018case}. Here, we identify the unobserved construct of interest $Y^c$ as whether a call to the hotline concerns a child \emph{at risk of harm}. As an observed proxy for risk, deployed predictive models estimate the probability of an \emph{out-of-home placement} of the child ($Y$). 
This indicator of risk is an important one, but there are many reasons why calls that merit screen-in may not represent a high risk for out-of-home placement. For instance, a case worker may be able to connect families to supportive services that would mitigate the risk of harm without family separation. 

The construct gap in this domain is easy to conceptually recognize, but its empirical implications have yet to be studied. In this empirical work, we first provide evidence showing that calls are received that likely warrant screen-in and that are being treated as high risk by child abuse hotline call workers, but the risk they represent is not reflected in $Y$ and therefore is not captured by models trained to predict this observed outcome. We do so by demonstrating that an out-of-home placement prediction model leads to poor performance with respect to other observable outcomes---outcomes that are not used as part of the predictive model for organizational reasons but that arguably capture relevant dimensions of risk.
We then demonstrate how models trained using our proposed methodology, which leverages historical expert consistency to augment observed outcomes, greatly improve precision on these other outcomes, while retaining comparable performance for out-of-home placement prediction and outperforming a model trained on human decisions with respect to this central target. 

\paragraph{Outline.}
In Section~\ref{sec:related} we review related work. In Section~\ref{sec:method} we describe the proposed approach to estimate expert consistency. We then introduce the label amalgamation strategy in Section~\ref{sec:method_lev}.  
In Section~\ref{sec:res} we empirically validate the methodology by conducting simulations on publicly available data followed by a set of experiments on real data from the child welfare context. Section~\ref{sec:disc} concludes with a discussion, managerial implications, and future work.

\section{Related work}
\label{sec:related}

Our work relates to different bodies of research and builds on previous work across various disciplines. In this section, we provide an overview of related work.  

\subsection{Machine learning for decision support}

Efforts to support expert decision making with algorithmic tools date back to the use of simple decision rules and statistical models~\citep{meehl1954clinical,sharda1988decision, dawes1989clinical,grove2000clinical} and have motivated the design of ML algorithms for decades~\citep{horvitz1988decision}. 
Challenges of designing and deploying these systems include dealing with societal biases encoded in data~\citep{barocas2016big, eubanks2018automating}, understanding how experts use algorithmic recommendations~\citep{tan2010assessing, lebovitz2019doubting, de2020case}, accounting for dynamic environments~\citep{meyer2014machine, d2020fairness,dai2021fair}, and assessing the effects of algorithms with respect to meaningful downstream endpoints~\citep{luca2016algorithms}. 

The challenge central to our work is the construct gap, which refers to a mismatch between an observed outcome $Y$ and an underlying construct of interest $Y^c$. Such a mismatch can arise as a result of different mechanisms. One such setting is when observed outcomes insufficiently capture the construct of interest, leading to an omitted payoff bias~\citep{chalfin2016productivity}. For instance, the objective when hiring may depend both on the productivity of an employee and on the length of the employee's tenure at the company; optimizing for only one of these factors results in an incorrect objective function.
More generally, mismeasured outcomes can result from imperfect proxies, which often is considered an issue of construct validity~\citep{mullainathan2017does, friedler2016possibility, chalfin2016productivity}. Constructs of interest often are complex, multi-faceted, and hard to quantify. This lack of a gold standard in observed outcomes arises in health care~\citep{adamson2019}, business~\citep{luca2016algorithms}, and criminal justice~\citep{fogliato2021validity}. 

\subsection{Learning from observational data of human decisions}

Human decisions and actions are used as training labels for many different purposes, ranging from inferring consumers' preferences~\citep{das2007, yoganarasimhan2020search} to imitation learning~\citep{argall2009survey, calinon2009robot}. A key difference between the tasks these studies consider and the task we consider is that we do not assume that humans constitute a ``gold standard." 
Far from being an oracle, expert decisions in high-stakes settings often involve high degrees of uncertainty, and in such cases models trained on human decisions can replicate and amplify human errors~\citep{mullainathan2017does}. For instance, predicting whether a child is at risk of maltreatment can be very difficult for human experts. Because some historical call-screening decisions may be erroneous, the goal in our context is not necessarily to imitate or automate the expert. 

Recent research in information systems has proposed new ways of extracting knowledge from historical expert decisions.~\citet{geva2019} propose the use of ML to assess experts' performance in the absence of gold standard labels. Although the overall objective of our work is significantly different, we share with~\citet{geva2019} the goal of using ML to leverage observational data that contains a single expert decision per case and no ground truth. Given the broad availability of this type of data in organizations' information systems, methodologies that facilitate knowledge extraction from historical data have the potential to improve organizational practices by enabling better decision making, and we contribute to this body of work.

\subsection{Learning from human consistency}

Our work draws inspiration from an extensive literature that uses inter-rater agreement metrics as an indicator of correctness~\citep{cohen1960coefficient,umesh1989interjudge, banerjee1999beyond}. Such metrics have been popular in applied psychology literature for decades, and they have been popularized in computer science through the crowdsourcing literature~\citep{snow2008cheap, amirkhani2014agreement}. With the emergence of an online workforce as an inexpensive source of data labeling~\citep{doan2011crowdsourcing}, metrics of agreement have been used to aggregate and assess the quality of crowd-sourced labels. In contrast to this crowdsourcing work, we aim to learn from domain experts' high-stakes decisions. Eliciting assessments by experts is not as viable a process as collecting labels from crowds. Such data often are sensitive, qualified labelers are scarce, and collecting multiple assessments for each case frequently requires setting up expensive review panels~\citep{gulshan2016development, mckinney2020international}. As a result, the incentive to find ways of leveraging observational data that contain historical expert decisions is strong. 
The proposed approach enables estimation of expert consistency from historical data even when only one expert has assessed each case. Here, we offer a caveat regarding consistency. Experts' consistency has been a subject of study for a long time \citep{shanteau1992competence, shanteau2015task}, and results generally have indicated that experts tend to exhibit low overall consistency. However, our approach depends not on overall consistency, but on consistency displayed on subsets of data. Our empirical results show that non-negligible subsets of consistently assessed cases do exist in real-world data.

\subsection{Design for human-ML collaboration}

Researchers have acknowledged that human-ML collaboration requires the design of an effective team and have explored several avenues to advance this goal. A stream of work has proposed a series of approaches to help the \emph{human} be a good team member. This work includes interventions to reduce both underreliance~\citep{dietvorst2018overcoming} and overreliance on the algorithm~\citep{buccinca2021trust}, to foster better use of information that is unobserved by the algorithm~\citep{holstein2023toward}, and to help humans to identify out-of-distribution samples~\citep{chiang2021you}.

Another stream of research that is more closely related to our work has focused on developing \emph{algorithms} that are better team members. For instance, researchers have proposed methodologies to prioritize features that are more relevant to experts~\citep{wang2018learning}. More closely related to our work, the literature on learning to defer~\citep{cortes2016learning,madras2017predict,wilder2020learning} has explored ways of dividing the labor between an algorithm and its human counterparts by allowing the algorithm to specialize in \emph{complementing} humans. 
Common to these approaches is the assumption that both the human and the algorithm are engaged in the same predictive task, which is well-captured by an observed target label $Y$. 

\subsection{Influence functions}

Finally, a core piece of related work is the literature on influence functions. The local influence method enables researchers to estimate the influence of a model's minor perturbations over a certain function, such as the loss or an individual prediction~\citep{cook1986assessment}. This fundamental work in the field of robust statistics has been widely applied in the literature on semi-parametric and nonparametric estimation~\citep{bickel1993efficient}, as well as causal inference~\citep{kennedy2016semiparametric}. In econometrics, influence estimation has been used to study how marginal perturbations of the data may change empirical conclusions derived from the application of standard econometrics methods~\citep{broderick2020automatic}. In ML,~\citet{koh2017understanding} have used influence functions 
as a way to explain black-box predictions and to generate adversarial attacks, and~\citet{koh2019accuracy} have studied the validity of influence functions to estimate the effect of removing a set of points from the training data. To the best of our knowledge, ours is the first work to propose the use of influence functions to estimate the influence of labelers or decision makers on model predictions.



\section{Estimation of expert consistency}
\label{sec:method}

In this section we propose a methodology to estimate expert consistency when only a single decision is observed for each instance. We then explain how we leverage the estimated expert consistency in Section~\ref{sec:method_lev}. 

Given an instance, we say that experts exhibit consistency if there is a decision that they are likely to make with high probability. 
Consistency can be directly estimated if multiple assessments are observed for each instance, using measures such as Cohen's kappa coefficient~\citep{cohen1960coefficient}.
However, in the absence of multiple available assessments per instance, consistency cannot be directly estimated.  
We propose an ML-based solution to solve this problem. In Section~\ref{sec:rand}, we formally introduce consistency and then tackle the problem relying on the assumption that the assignment of experts-to-instances is random. However, this assumption frequently is violated in practice. Thus, in Section~\ref{sec:inf}, we extend our approach to avoid this assumption and estimate expert consistency when the assignment of experts-to-instances is non-random.

\subsection{Indirect estimation of expert consistency}
\label{sec:rand}

Let $X$, $D$, and $H$ denote random variables corresponding to the covariates, the observed binary human decision, and the identifier of the expert who made the decision, respectively. Lowercase letters denote the realisation of these variables. We assume that each instance $i$ is represented by a set of observed features $\bm{x}_i$ and has an associated decision $d_i$ made by expert $h_i\in \{1,...,q\}$, where $q$ is the number of experts. For notation clarity, we use the subscript $i$ when dependencies exist between the variables' realisations and omit it otherwise. Let $\mathbf{X}\in \mathbb{R}^{m \times n}$ denote the matrix of all training data, where \textit{m} is the number of instances and \textit{n} is the number of features.  Table~\ref{table:notation} presents a summary of the notations.

Formally, we say that experts exhibit $\delta$-level consistency for a given instance if there is a decision $d$ such that the probability that a randomly selected expert will make decision $d$ is greater or equal to $\delta$.
When multiple decisions are observed for each instance, directly estimating consistency is easy. For example, if 10 experts assess a given instance, then we estimate a 0.8-level of consistency if at least 8 of the experts' decisions are the same. 
Our goal is to estimate expert consistency when we observe only one expert decision for each instance. 

If the assignment of experts-to-instances is independent of the covariates (i.e., $X\ind H$), we can formally express $\delta$-level consistency for an instance $\bm{x}$ as:
\begin{equation}
P(D=d \mid X=\bm{x}) \ge \delta.
\label{eq:delta_cons}
 \end{equation}
As we prove in the derivation included in Appendix~\ref{app:consistency}, this quantity does not depend on which expert assessed each case. If the instances in $\mathbf{X}$ are sampled i.i.d. from the same distribution, we can estimate this quantity via a machine learning model. 
Let ${f}_D$ denote a predictive model of expert decisions, ${f}_D={P}(D=1 \mid X)$, where ${P}$ corresponds to the estimated probability. Note that, consistent with Eq.~\eqref{eq:delta_cons} and the proof in Appendix~\eqref{app:consistency}, this is a single model that predicts observed expert decisions, without requiring information of which expert assessed each instance. If ${f}_D$ is well-calibrated\footnote{Calibration can be empirically assessed. Some models, such as logistic regression, are inherently well-calibrated. When an empirical assessment indicates that the estimated probabilities are not well-calibrated, calibration techniques, such as Platt's scaling~\citep{platt1999probabilistic}, can be applied. In such cases, our methodology inherits any requirements of the chosen recalibration methods. We do note, however, that these restrictions typically are not too limiting. Most models benefit from Platt recalibration if their predictions are sufficiently ordered~\citep{niculescu2005predicting,caruana2006empirical}, which can be assessed via the Area Under the ROC Curve (AUC).}, meaning that the model's predicted scores align with the probabilities of the observed decisions, the set $\mathcal{A}_{rand}$ defined in Eq.~\eqref{eq:A_def} corresponds to the cases for which a certain decision is predicted with a probability greater than or equal to $\delta$. This set constitutes an \emph{estimated} $\delta$-level consistency set. For simplicity, we refer to it as $\mathcal{A}$ throughout the text, but note that the variant below is suitable when assignment of experts to cases is random. In Section~\ref{sec:inf} we define a variant that is suitable for non-random assignment. 
 \begin{eqnarray}
\mathcal{A}_{rand}:= \{\bm{x}_i \in \mathbf{X} : |{f}_D(\bm{x}_i) - d_i| \geq \delta \} = \{\bm{x} \in \mathbf{X} : ({f}_D(\bm{x}) \geq \delta) \lor ({f}_D(\bm{x}) \leq 1 - \delta)\}
\label{eq:A_def}
 \end{eqnarray}
In Theorem~\ref{theorem:mis} below, we show that this 
is a good estimate of a $\delta$-level consistency set. Before we show the formal result, we provide the intuition for it. First, note that we make a one-directional claim: if an instance $\bm{x}\in \mathcal{A}$, then $\bm{x}$ has $\delta$-level consistency. However, $\mathcal{A}$ may not contain \emph{all} instances for which experts exhibit $\delta$-level consistency.
The reason is that modeling choices affect our ability to accurately predict expert decisions. In particular, model misspecification may result in a model that is uncertain about experts' decisions for a certain subset (e.g., estimated $0.5$ probability of $D=1$ for an instance $\bm{x}$), even if experts consistently make the same decision. 
However, when we estimate that an instance lies in the $\delta$-level consistency set, the estimation does not rely on functional form assumptions. That is, the choice of model used to estimate ${f}_D$ (e.g., a logistic regression) does not need to correspond to the true relationship between $X$ and $D$. 
As an illustrative example, consider a 0.95-level consistency set. This estimate means that a well-calibrated model predicts that for instances in this set, a certain decision will be made 95\% of the time. Thus, the model chosen for ${f}_D$ was a good choice for predicting decisions for this subset of instances because calibration implies that humans indeed make the same decision for 95\% of instances in this set.

The fact that $\mathcal{A}$ is a good estimate of a $\delta$-level consistency set can be formally proved by estimating the confidence interval of the true probabilities that we care about (i.e., $P(D=1 \mid X)\geq \delta$ and $P(D=1 \mid X)\leq 1-\delta$) for instances in the set $\mathcal{A}$. 
Note that the set $\mathcal{A}$ can be expressed as $\mathcal{A}=\mathcal{A}_{0}\cup \mathcal{A}_{1}$, where $\mathcal{A}_{0}$ contains high-probability predictions of $D=0$, and $\mathcal{A}_{1}$ contains high-probability predictions of $D=1$.

\begin{theorem}
\label{theorem:mis}
Given an estimated probability ${f}_D={P}(D=1 \mid X)$ and a set $\mathcal{A}_{1}=\{\bm{x} \in \mathbf{X}  : {f}_D(\bm{x})\geq \delta \}$, the confidence interval of the true probability, $P(D=1 \mid X)$, for data points in $\mathcal{A}_{1}$ can be estimated as:
 \begin{eqnarray}
    \mathsf{CI}(P(D=1 \mid X=\bm{x}, \bm{x}\in \mathcal{A}_{1})\geq \delta \hspace{0.01in} ;\hspace{0.01in} C) = \left(   \delta- z^{\ast}\frac{\sigma}{\sqrt{k}} ,  1 \right) , 
    \label{eq:ci}
\end{eqnarray}
where $CI(P; C)= (a,b)$ denotes that $(a,b)$ is the confidence interval of P at level C. Here, $\mathbf{X}_{v}\in \mathbb{R}^{k \times n}$ is a validation set drawn from the same distribution as $\mathbf{X}$; $k$ is the number of instances in this validation set; 
$\sigma$ is the standard deviation of the set $\{ d_i  \hspace{0.02in}   \mid   \hspace{0.02in}  \bm{x}_i\in \mathbf{X}_{v}, {f}_D(\bm{x}_i) \geq \delta\} $, which contains the expert decisions for instances in the set $\mathbf{X}_{v}$ that are estimated to lie in set $\mathcal{A}_{1}$; and $z^{\ast} = \Phi^{-1}(1-\frac{\alpha}{2})$ for $\alpha = \frac{1-C}{2}$ and $\Phi$ the standard normal distribution. (Proof is in Appendix \ref{proof:mis}.)

\end{theorem}

This theorem says that, for instances in the set $\mathcal{A}_{1}$, the C confidence interval of the true probability $P(D=1|X)$ is lower bounded by $\delta- z^{\ast}\frac{\sigma}{\sqrt{k}}$. An analogous proof holds for $\mathcal{A}_{0}$. Note that $\lim_{k \to\infty} (z^{\ast}\frac{\sigma}{\sqrt{k}})=0$, in which case the lower bound converges to $\delta$. Furthermore, when $\delta$ is chosen such that it constrains $\mathcal{A}$ to high-probability predictions (i.e., the value of $\delta$ is large), the standard deviation $\sigma$ is small. 
For example, assume that we are estimating the $0.9$-level consistency set---that is, the set of instances for which we estimate that experts are likely to make a certain decision with a probability of 90\%. Furthermore, assume that in this given example, the validation set is of modest size $k=5000$ and, as was previously assumed, that ${f}_D$ is well-calibrated. Then, the corresponding 95\% confidence interval is lower bounded by $0.892$, a remarkably tight bound.\footnote{$z^{\ast}$ of a 95\% confidence interval is 1.96; assuming a well-calibrated model $\sigma=0.2\overline{9}$; thus, $\delta- z^{\ast}\frac{\sigma}{\sqrt{k}}= 0.9-1.96*(0.3/\sqrt{5000}) =0.892$.} Thus, our theoretical result shows that the proposed approach yields a good estimation of expert consistency and that the set $\mathcal{A}$ indeed contains instances for which experts exhibit consistency at the estimated level.

\subsection{Expert consistency estimation under non-random assignments}
\label{sec:inf}

The expert consistency estimation methodology proposed in Section~\ref{sec:rand} assumes that the assignment of experts to cases is random. This assumption is plausible in some settings, but it is unlikely in others. For example, in the context of child abuse hotlines, call workers who routinely work on weekends or at night may observe different cases than those who primarily work on weekdays.  
In this section, we extend our methodology to address the non-random assignment problem. 

When assignment of experts-to-instances is non-random, a risk arises that high-probability predictions of human decisions, obtained by ${f}_D$, are primarily driven by the decisions of a single expert. For example, if one expert observed all instances of a certain type and always made the same decision, then the model would estimate high consistency for such instances. Thus, under non-random assignment, the set defined in Eq.~\eqref{eq:A_def} may not reflect consistency across experts but instead may 
reflect idiosyncrasies or individual biases.

To address this problem, we propose using the local influence method, a technique from robust statistics that assesses whether a statistical result of interest is sensitive to minor perturbations of a model~\citep{cook1986assessment}. One such family of perturbations involves small disturbances in the data used to train a predictive model. In this case, the local influence method can be used to assess how predictions would change if the model were trained on a slightly different training dataset. 

Recall that we estimate consistency using a machine learning model ${f}_D={P}(D=1 \mid X)$, which we have shown to be a good mechanism for estimating consistency in the case of random assignments (i.e., $X \ind H$). In the case where $X \not\ind H$, we propose to estimate how perturbing the data---by marginally upweighting the instances an expert assesses---would affect the predictions made by ${f}_D$. By doing this estimation for each expert, we can assess whether the estimate of interest, ${P}(D=d \mid X =\bm{x})> \delta$, is robust to perturbations related to the importance given to each expert. Intuitively, this work allows us to discard instances for which the estimated consistency is too sensitive to one or a very few experts. 

We describe how to estimate the influence of each expert in Section~\ref{subsec:inf}. After we have in place the machinery to estimate each expert's influence, we propose in Section~\ref{subsec:consistency} to constrain the high-consistency set $\mathcal{A}$ only to the instances $\bm{x}$ for which the predicted probability ${f}_D(\bm{x})$ is both within the desired range and robust to perturbations of importance granted to individual experts.

\subsubsection{Influence of a single expert}
\label{subsec:inf}

Consider a weight vector $\bm{w}\in \mathbb{R}^m$ that represents the weights associated with the training instances in $\mathbf{X}^{m\times n}$. The default training mode in ML is to give all instances equal weight: $\bm{w} = [1,1,...,1]$.
We can perturb the data in a direction corresponding to an expert $h$ by defining a corresponding weight $\bm{w}^h\in \mathbb{R}^m$.
Recall that $h_i \in \{1,...,q\}$ indicates the expert who assessed the instance $i$ with covariates $\bm{x}_i \in \mathbf{X}$. Given a decision-maker $h$, let the vector $\bm{w}^h \in \mathbb{R}^{m}$ be defined as: 
\begin{equation}
\begin{array}{rcl}
\forall i \in [\![1, m]\!], \bm{w}^h_i & := & \left\{ \begin{array}{rcl}
1 + \varepsilon & \mbox{for} & h_i = h \\
1 & \mbox{for} & h_i \neq h  \\
\end{array}\right. .
\end{array}
\label{eq:w}
\end{equation}

A perturbation of the data in the direction $\bm{w}^h$ corresponds to an increase in the importance of expert $h$ by up-weighting the instances assessed by this expert by an infinitesimal amount, $\varepsilon \in \mathbb{R}^+$. The function $\mathcal{I}_{D}(\bm{w}^h,\bm{x})$, defined in Eq.~\eqref{eq:inf_norm}, estimates the normalized influence on the predicted probability ${f}_D(\bm{x})$ of this perturbation.
In other words, this influence function allows us to answer the question: if we up-weight the importance given to expert $h$ by $\epsilon$, how will the predicted probability ${f}_D$ for the instance $\bm{x}$ change? 

The influence of expert $h$ on a prediction can be derived in analogous fashion to the way~\cite{koh2017understanding} derive the influence of perturbing a single data point by $\varepsilon$. Following the notion of local influence introduced in~\cite{cook1986assessment}, we can define the influence of perturbing the data by $\bm{w}^h$ in terms of $\epsilon$, as specified in Eq.~\eqref{eq:infl_p_def},
\begin{equation}
\begin{array}{lll}
 \left. \frac{\partial {P}(d \mid \bm{x},\hat{\theta}_h)} { \partial \epsilon}\right \rvert_{\varepsilon=0} .
\end{array}
   \label{eq:infl_p_def}
\end{equation}

Here, the empirical risk minimizer is $\hat{\theta}:= \textrm{argmin}_{\theta \in \Theta}\sum_{i=1}^n \mathcal{L}(\bm{x}_i,d_i,\theta)$, and the empirical risk minimizer after the training data has been perturbed by $\bm{w}^h$ is $\hat{\theta}_h:= \textrm{argmin}_{\theta \in \Theta}\sum_{i=1}^n w^h_i \mathcal{L}(\bm{x}_i,d_i,\theta)$. $\mathcal{L}(\bm{x}_i,d_i,\theta)$ is the loss function, and $\theta$ denotes the model parameters, $\theta \in \Theta$, where $\Theta$ is the parameter space. When comparing the influence of multiple experts, we may want to account for the fact that some experts may have observed more cases than others; thus, we want to consider the \emph{normalized} influence, where the instances in the training set assessed by expert $h$ are $ \mathbf{X}_h = \{\bm{x}_i \in \mathbf{X}  \mid  h_i = h\}$. Thus, the normalized influence is defined in Eq.~\eqref{eq:inf_norm}. 
\begin{equation}
\label{eq:inf_norm}
   \mathcal{I}_{D}(\bm{w}^h,\bm{x}) :=  \left. \frac{1}{| \mathbf{X}_h|} \frac{\partial {P}(d  \mid  \bm{x},\hat{\theta}_h)} { \partial \varepsilon}\right \rvert_{\varepsilon=0} 
\end{equation}
  
The quantity in Eq.~\eqref{eq:inf_norm} is intuitive---it captures how fast the predicted probability changes when the weight given to an expert during training varies. However, this quantity cannot be directly estimated as the relation between ${P}$ and $\varepsilon$ cannot in general be expressed in a closed form. 
The following Theorem shows how the influence function can be approximated in estimable terms. 

\begin{theorem}
\label{theorem:inf}
Assuming a doubly differentiable loss $\mathcal{L}$ and an invertible Hessian,
the normalized influence $\mathcal{I}_{D}(\bm{w}^h,\bm{x})$ of perturbing the training data $\mathbf{X}\in \mathbb{R}^{n\times m}$ in the direction $\bm{w}^h\in \mathbb{R}^{n}$ can be approximated as:
\begin{equation}
\label{eq:influence}
 \mathcal{I}_{D}(\bm{w}^h,\bm{x})  \approx  -\frac{1}{| \mathbf{X}_h|} \nabla_{\theta}{P}(d \mid \bm{x},\hat{\theta})^T [\nabla^2_{\theta}\mathcal{R}(\hat{\theta})]^{-1}
 \left[\sum_{i,\bm{x}_i \in  \mathbf{X}_h} \nabla_{\theta} \mathcal{L}(\bm{x}_i,d_i,\hat{\theta})\right],
\end{equation}

where $\nabla_{\theta}$ denotes a gradient with respect to $\theta$ and where $\mathcal{R}(\theta):= \sum_{i=1}^n \mathcal{L}(\bm{x}_i,d_i,\theta)$ is the empirical risk.  (See the proof in Appendix \ref{proof:inf}.)
\end{theorem}

Now the influence is defined in terms of $\theta$ instead of $\varepsilon$ and $\theta_h$ and therefore can be easily estimated. Since $\theta$ is a function of the training data points, we recommend employing cross-validation to assess the influence of each training point. This separation of the estimation of influence and $\theta$ effectively mitigates the potential risk of information leakage. Finally, the most computationally intensive component is inverting the Hessian of the empirical risk, but approaches to approximate it efficiently for complex models have been proposed~\citep{koh2017understanding}. 

In addition, note that the assumptions made in Theorem~\ref{theorem:inf} are driven by the need to invert the Hessian to estimate the influence. These assumptions carry practical implications. 
Because the loss function must be doubly differentiable with respect to the model's parameters, the methodology can easily be applied not only when using simple methods, such as logistic regression, but also when using complex methods, such as deep neural networks. However, it cannot be (directly) applied to some methods, such as tree-based algorithms, because it would require an approximation of the gradient of the loss. 
Whether the Hessian is invertible is also data-dependent. This condition may be violated under certain settings, such as when strong collinearity exists across features. Fortunately, this does not constrain the datasets to which the method can be applied. Instead, in such cases, approaches such as data pre-processing via feature selection or dimensionality reduction can help transform the feature space into one that meets the necessary conditions. Alternatively, loss regularization, such as the addition of an $L_1$ penalty, can address the issue. 

\subsubsection{Estimating consistency}
\label{subsec:consistency}

After the influence of each expert is estimated, we can determine whether the predictions of the model ${f}_D$ are robust to perturbations on the weights given to individual experts. 

For a given data point $\bm{x}$, we can analyze the distribution over the influence functions $ \mathcal{I}_{D}(\bm{w}^h,\bm{x})$, $\forall h\in \{1,...,q\}$. 
Let $\bm{s}(\bm{x})$ be a vector of the absolute influence of each decision maker over ${f}_D(\bm{x})$, sorted in decreasing order, where $s_j(\bm{x})$ denotes the $j^{th}$ entry of $\bm{s}(\bm{x})$. Formally, we define: 
\begin{equation}
\bm{s}(\bm{x}):=sort([|\mathcal{I}_{D}(\bm{w}^h,\bm{x})|\textrm{, for }h\in\{1,...,q\}]).
\label{eq:s}
\end{equation}

The task now is to determine whether a distribution of influence is indicative of expert consistency. We propose the following three metrics as heuristics to assess this consistency.

\textbf{Center of mass.} The center of mass of influence measures whether influence is spread across experts or whether very few of the experts have a disproportionate influence on a prediction ${f}_D(\bm{x})$. 
The center of mass $m_1(\bm{x})$, defined in Eq.~\eqref{eq:m1}, indicates the number of experts who together have absolute cumulative influence matching the influence of the rest of the experts for instance $\bm{x}$.
For instance, a center of mass of $m_1(\bm{x})=5$ indicates that the $5$ experts with the greatest influence have as much influence as all the rest of the experts. A small center of mass indicates one or a few experts have a disproportionate influence.
\begin{equation}
    m_1(\bm{x}):=\frac{\sum_{j}  j \cdot s_j(\bm{x})}{\sum_j  s_j(\bm{x})} 
    \label{eq:m1}
\end{equation}

\textbf{Aligned influence.} The center of mass captures the \textit{concentration of influence}, but it does not take into account directionality. A positive influence, $\mathcal{I}_{D}(\bm{w}^h,\bm{x})>0$, means that the predicted probability ${f}_D(\bm{x})$ would increase if the training data is perturbed by $\bm{w}^h$. Meanwhile, a negative influence means that the predicted probability would be reduced under the specified perturbation. 
To incorporate this element, the second metric, $m_2(\bm{x})$, reflects the extent of \textit{opposing influences} and is defined in Eq.~\eqref{eq:m2}. This metric indicates the portion of influence going in the direction of the predicted probability; that is, if ${f}_D(\bm{x})\geq 0.5$, the numerator consists of the sum of positive influences, and if ${f}_D(\bm{x})<0.5$, the numerator consists of the sum of negative influences. Note that if all experts drive the prediction in the same direction, then $m_2(\bm{x}) =1$.
\begin{equation}
\begin{array}{rcl}
m_2 (\bm{x}) & := & \left\{ \begin{array}{rl}
    \frac{\sum_{i, \mathcal{I}_{D}(\bm{w}^i,\bm{x}) > 0} s_i(\bm{x})}{\sum_i s_i(\bm{x})} & \mbox{if } {f}_D(\bm{x}) \geq 0.5 \\
    \frac{\sum_{i, \mathcal{I}_{D}(\bm{w}^i,\bm{x}) < 0} s_i(\bm{x})}{\sum_i s_i(\bm{x})} & \mbox{if } {f}_D(\bm{x}) < 0.5 \\
    \end{array}\right.
\end{array}
\label{eq:m2}
\end{equation}

\textbf{Negligible influence.} 
When no perturbation over expert weights would influence a prediction, we have a strong indication of consistency. This situation is captured by the maximum influence among experts, as defined in Eq.~\eqref{eq:m3}. Given an arbitrarily small $\gamma_3$, if $m_3(\bm{x})<\gamma_3$, we can infer consistency, given that no perturbation would significantly influence the prediction.
\begin{equation}
    m_3(\bm{x}) := \mbox{max}(s(\bm{x}))
    \label{eq:m3}
\end{equation}
These metrics can be used to constrain the high consistency set $\mathcal{A}$ to instances where the predictions are robust to perturbations on the weight given to experts. Specifically, we can constrain the set to instances where either all experts' influence is negligible \emph{or} where there is both a large center of mass and well-aligned influences. This allows us to estimate consistency when there is non-random assignment of experts to cases. Given parameters $\gamma_1,\gamma_2,\gamma_3 \in \mathbb{R}^+$, we define the set $\mathcal{A}_{non-rand}$ as,
\begin{equation}
\resizebox{.92\hsize}{!}{$
    \mathcal{A}_{non-rand}:=\left\{\bm{x}_i \in X  : [|{f}_D(\bm{x}_i)-d_i| < \delta] \wedge \left[\big([m_1(\bm{x}_i)>\gamma_1] \wedge [m_2(\bm{x}_i)>\gamma_2] \big) \lor [m_3(\bm{x}_i) < \gamma_3] \right] \right\}.$}\label{eq:A_c}
\end{equation}

The metric $m_3(\bm{x})$ has the advantage of not requiring any type of parameter tuning because $\gamma_3$ is an arbitrarily small parameter that can be used off-the-shelf. Thus, we have a conservative, off-the-shelf starting point for consistency estimation, setting $ \gamma_1=\lfloor q/2 \rfloor$ and $\gamma_2=1$---their maximum possible values. Users can then select other values for $ \gamma_1$ and $\gamma_2$ if they deem more amalgamation to be appropriate. For example, setting $\gamma_2=0.9$ indicates that no more than 10\% of the influence may be going in a direction counter to the prediction. 

All three metrics offer the advantage of having a clear conceptual grounding and interpretation, facilitating the process of choosing parameters and interpreting results. In Section~\ref{subsec:semi}, we also show empirically that they serve to correctly estimate consistency in the presence of a non-random experts-to-instances assignment. However, it is possible that additional useful metrics and heuristics may also be derived from the distribution of experts' influence, which may serve for consistency estimation, as well as for other tasks. We hope that the proposed methodology to estimate experts' influence can serve as a building block for others who are studying expert decision-making and human-AI collaboration.

\section{Leveraging estimated expert consistency}
\label{sec:method_lev}
Once consistency across experts has been estimated through Eq.~\eqref{eq:A_def} or Eq.~\eqref{eq:A_c}, depending on the expert assignment regime, the inferred expert knowledge can be leveraged to improve algorithmic decision support. Specifically, we can use the set of instances for which experts are likely to agree in their assessment, denoted in the remainder as $\mathcal{A}$, to reduce the construct gap.
In Section~\ref{sec:augm}, we propose a label amalgamation approach that incorporates inferred expert consistency into the labels used to train an ML model. 

\subsection{Label amalgamation}
\label{sec:augm}

 The proposed label amalgamation approach provides a way to simultaneously learn from expert decisions, $D$, and from observed outcomes, $Y$. The label amalgamation defines a label $Y^{\mathcal{A}}$ such that, for every instance $i$, this label is equal to the observed human decision $d_i$ if experts are estimated to be consistent for this instance, and it defaults to the observed outcome $y_i$ otherwise\footnote{The amalgamation can be done on the entire set $\mathcal{A}$, as considered throughout this article, or asymmetrically, amalgamating only one of two possible decisions. Appendix~\ref{app:asymmetric} describes asymmetric amalgamation.}.
Recall that $\mathcal{A}$ is the inferred $\delta$-level consistency set. 
Formally, we define the vector of amalgamated labels $Y^{\mathcal{A}}$ as
\begin{equation}
\begin{array}{rcl}
\forall i \in [\![1, m]\!], Y^{\mathcal{A}}_i & = & \left\{ \begin{array}{rcl}
d_i & \mbox{if } & \bm{x}_i \in \mathcal{A} \\
y_i & \mbox{if } & \bm{x}_i \notin \mathcal{A} \\
\end{array} .\right.
\end{array}
\label{eq:amalg}
\end{equation}

As an example, consider the child abuse hotline domain. As summarized in Table~\ref{tbl:examples}, the decision $D$ corresponds to the call workers' decision to screen in a call for investigation, and the observed outcome, $Y$, corresponds to out-of-home placement of the child. In this context, the set $\mathcal{A}$ corresponds to the cases for which the estimation indicates that call workers consistently make the same screen-in decision. Thus, the amalgamated label comprises two elements: the call workers' decision for instances in $\mathcal{A}$, and the observed outcome for all other instances.  

As previously stated, the core assumption of our work is that expert consistency is indicative of correctness. Formally, this corresponds to assuming that $\forall x_i \in \mathcal{A}, P(Y^c= d_i \big| X= \bm{x}_i) \geq P(Y^c=y_i \big|X = \bm{x}_i)$. The conceptual motivation for choosing to infer and amalgamate experts' consistency is not much different from the motivation to assemble expert panels for high-quality data labeling; both hinge on a contextual understanding that justifies the belief that expert consistency encodes knowledge. Meanwhile, we assume that in cases where there is no estimated consistency, a decision coming from a single expert is less reliable, since it may reflect idiosyncrasies or biases. Thus, when we do not infer high expert consistency we default to learning from the observed outcome. Formally, this corresponds to assuming that $\forall \bm{x}_i \not\in\mathcal{A}, P(Y^c=y_i \big| X = \bm{x}_i) \geq P(Y^c=d_i \big| X = \bm{x}_i)$. Under such assumptions, it naturally follows that $\mathbb{E}(|Y^c-Y^{\mathcal{A}}| \big| X)\leq \mathbb{E}(|Y^c-Y| \big| X)$ and $\mathbb{E}(|Y^c-Y^{\mathcal{A}}| \big| X)\leq \mathbb{E}(|Y^c-D| \big| X)$.
This indicates that label amalgamation, which allows us to then train a model to estimate $\mathbb{E}(Y^{\mathcal{A}} \big| X)$, provides a good path to training a decision support system under these assumptions. 




Thus, having defined the amalgamated label $Y^{\mathcal{A}}$, we can train a predictive model to simultaneously learn from inferred expert consistency and observed outcomes:
\begin{equation}
    {f}_{\mathcal{A}}(\bm{x})={P}(Y^{\mathcal{A}} | X=\bm{x}).
\end{equation}
Importantly, note the enhanced flexibility in our proposed methodology using amalgamated labels: The model family used for estimating consistency---which has some constraints (e.g., being doubly-differentiable)---needs not be the same one used for training the model.

\begin{figure}[h]
\centering
\includegraphics[width=\linewidth,trim={0 8cm 0 9cm},clip]{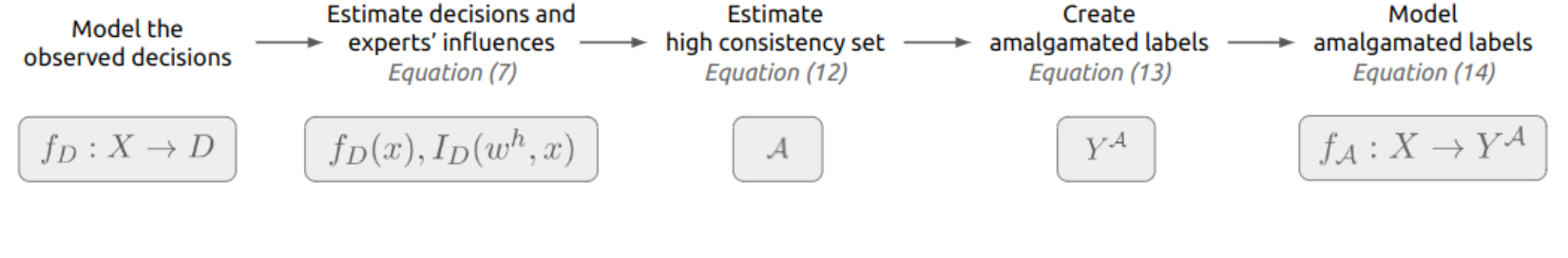}
\caption{Diagram summarizing the steps of the proposed methodology.}
\label{fig:summary}
\end{figure}

Figure~\ref{fig:summary} summarizes the steps to apply the proposed methodology. We also provide a step-by-step Python tutorial (described in an algorithmic form in Appendix~\ref{app:algo_summary} and available on Github\textsuperscript{4}) for executing this pipeline on a publicly available dataset, incorporating all practical recommendations for model calibration and cross-validated influence estimation. The accompanying code is designed for seamless adaptation, allowing practitioners to adjust it to their specific dataset and expertise in estimating the amalgamation set.

\section{Analysis and results}
\label{sec:res}

In this section, we validate the proposed methodology using a simulated dataset based on the publicly available Medical Information Mart for Intensive Care--Emergency Department (MIMIC-ED) dataset~\citep{johnson2021mimic}. We use simulation strategies to assess the method's performance under different scenarios of human expertise and biases. We then present the main empirical findings of our work for the child welfare context---a high-stakes domain in which algorithmic decision support tools are increasingly being used~\citep{saxena2020human}. To ensure reproducibility and facilitate practitioners' use of the proposed approach, we have made all our experiments on MIMIC-ED and methodology available on GitHub\footnote{Github repository: \url{https://github.com/mariadea/influence-labelers}.}.  

\subsection{Experiment Setup}
\label{subsec:exp}
In all experiments, we compare performance across six baselines in addition to the proposed approach. 
The assessed models, and the rationale for using them as baselines, 
are the following:

\begin{itemize}
    \item ${f}_Y$: Model trained to predict an observed outcome, $Y$. Across many domains, such as the child welfare setting we study, solely using $Y$ is frequently the preferred choice.
    \item ${f}_D$: Model trained to predict human decisions, $D$.  This corresponds to attempting to ``automate" experts.
    \item ${f}_{noise}$: Model trained to robustly learn from noisy labels, $D$, via confident learning~\citep{northcutt2017learning}. Given that our motivation is that human decisions encode knowledge but may be noisy or biased, a model that learns from noisy labels is a natural baseline. Crucially, models that learn from noisy labels must make assumptions about the form of the noise, and many such models assume noise to be random. We chose \citet{northcutt2017learning} because it is more flexible in its assumption of noise form, and allows noise to depend on the class.
    \item ${f}_{def}$: Deferral model proposed by \citet{madras2017predict}. 
    This is an approach that also aims to leverage $D$ and $Y$, with the goal of improving performance via human-AI complementarity. The model simultaneously learns a classifier and a router: The classifier specializes on instances most difficult to humans, while the router determines which instances to route to the classifier and which to a human expert. It assumes $Y$ is a perfect label. 
    \item ${f}_{ens}$: Ensemble model. A model that trains two separate classifiers to predict $Y$ and $D$ and then averages their output. This is a standard approach of blending two sources of labels. 
    \item ${f}_{weak}$: Weak supervision model. Following the programmatic weak supervision paradigm, $Y$ and $D$ are treated as imperfect sources of labels. Only two sources of weak labels are available, thus an equal weighted vote is used as the label model (see e.g. \cite{ratner2017snorkel}). 
    \item ${f}_\mathcal{A}$: Proposed model trained on amalgamated labels, as defined in Equation~\eqref{eq:amalg}.
\end{itemize}


In addition to the baselines, we perform ablation studies to assess the value of the different components of our method; the results presented in Appendix~\ref{app:ablation} indicate that both the influence estimation and the label amalgamation contribute to the method's performance and robustness. 

For all models, we use multilayer perceptrons with ReLU activation functions. For details on training, see Appendix~\ref{app:training_det}.

\subsection{Decision-making simulations}
\label{subsec:semi}

In the first set of experiments, we study the performance of the model in simulations, using a publicly available dataset. Consider the tasks of triaging or screening patients for priority care~\citep{caruana2015intelligible,hong2018predicting,raita2019emergency}. 
Models that aim to support triaging decisions by predicting observed outcomes, such as hospitalization or medical complications, may fail to identify patients whose needs are not reflected in these outcomes or for whom appropriate prioritization resulted in a positive outcome~\citep{caruana2015intelligible}. In this section, we use this motivating example to study the methods' performance in controlled simulations. 

Suppose that a triaging task seeks to prioritize patients based on their needs, and that hospitalization is available as an observed outcome, $Y$, to train a decision support tool. However, 
not all patients that require prioritization are hospitalized, because ambulatory care may be urgent but sufficient. This means that $Y^c = Y \lor Y^{omitted}$, where $\lor$ is the binary logic operator \emph{or}. That is, $Y^c = 1$ if $Y=1$ or $Y^{omitted}=1$, where $Y$ is hospitalization and $Y^{omitted}$ is an alternative type of risk that does not lead to hospitalization but requires prioritization. Across different simulations, we assume that triage experts take these factors into consideration, with varying degrees of accuracy and bias. 


Using the MIMIC-ED dataset~\citep{johnson2021mimic}, we use vital signs measured at admission to the emergency room as covariates $X$, and we simulate different scenarios of the potential relationships between an observed outcome $Y$, a construct of interest $Y^c$, and human decisions $D$. The simulation allows us to use real covariates, while maintaining full control over the relationships between $Y$, $Y^c$ and $D$. We simulate $Y$, $Y^c$ and $D$ using tree-based structures trained to model the relationship between the covariates $X$ and outcomes observed in the data. We include all details of our simulation in Appendix~\ref{app:mimic_generation}. 
Throughout our simulations, we assume that there are 20 triage experts who assess cases, i.e. $q = 20$. Unless otherwise stated, we assume a random experts-to-cases assignment. 
In this set of experiments, parameters for expert consistency estimation were fixed to conservative values $(\delta, \gamma_1, \gamma_2, \gamma_3) = (0.05, 6, 0.95, 0.002)$; further sensitivity analyses are included in Appendix~\ref{app:mimic_hyperparam}. Motivated by the real-world dataset we study in Section~\ref{subsec:res_child}, we report analogous results to all simulations assuming a selective labels problem in Appendix \ref{app:mimic_selective}. This allows us to study the performance of the methods when $Y$ is only observed conditioned on $D$, and thus only one proxy is available whenever $D=1$.

\subsubsection{Learning from imperfect experts}\label{subsub:mimic_scenarios} We first study a set of scenarios that consider different degrees and types of erroneous beliefs. These scenarios consider the heterogeneity across experts, i.e., whether they are all equally likely to make a mistake for a given case, and the error frequency. In particular, we explore the following scenarios, representing different relationship between simulated decisions $D$ and simulated labels $Y$ and $Y^c$ (more details on the simulation for each of these scenarios can be found in Appendix~\ref{app:mimic_generation}):

\begin{itemize}
    \item \emph{Correct homogenous beliefs and randomness:} In this setting, decisions are either random or highly accurate. We assume that experts are bad at assessing patients who will be hospitalized and who do not present markers of the omitted risk. Specifically, we assume they make random decisions when $Y=1$ and $Y^{omitted}=0$. We assume that they can predict with very high accuracy the needs of all other patients, even if the patients may not require hospitalization.
    \item \emph{Correct and incorrect homogenous beliefs:} In this setting, incorrect decisions are no longer random. We assume that experts are systematically bad at assessing patients who will be hospitalized and who do not present markers of the omitted risk, making the wrong decision with regard to hospitalisation in approximately 75\% of these cases. We assume that they can predict with very high accuracy the needs of all other patients, even if the patients may not require hospitalization.
    \item \emph{Correct and incorrect heterogeneous beliefs:} In this setting, we assume that not all experts are equally likely to make a correct decision, and instead, their performance varies when assessing patients who will be hospitalized and who do not present markers of the omitted risk. For each expert, the probability of error with respect to hospitalization in these cases is $p_h$, for $p_h \sim \mathcal{U}(a,b)$. We assume they can all predict with very high accuracy the needs of all other patients, even if the patients may not require hospitalization. We present results for $p_h \sim \mathcal{U}(0.3,0.6)$ and consider another two variations of this setting with different $a, b$ parameters in Appendix~\ref{app:corr_incorr}.
    \end{itemize}

Table~\ref{tab:sim_auc_err} shows results under these different scenarios, reporting Area Under the ROC Curve (AUC) with respect to $Y^c$, which is known in the simulations. Across all scenarios, label amalgamation yields the best performance---a significant improvement compared to all other alternatives. The benefits of the proposed methodology are especially evident in comparison to ${f}_D$; even in settings where a model trained on expert assessments alone would yield worse than random performance, the proposed methodology is able to extract valuable information contained in experts' decisions to obtain substantial performance improvements with respect to all alternatives. As expected, the benefits of leveraging estimated consistency are especially salient for those patients whose risk is poorly proxied by hospitalization. Consider the first scenario, \textit{Correct homogenous beliefs and randomness};  Among 70,198 patient visits in the data, if the prioritization threshold was set to 70\% (as observed in the real data), ${f}_A$ would correctly prioritize an average of 8,446.4 patients whose risk is not captured by $Y$.


Importantly, the proposed methodology is the only one that is able to simultaneously outperform ${f}_Y$ and $f_D$ (except for one setting in which $f_{noise}$ outperforms both by a very small margin). Note that $f_{def}$ and $f_{\mathcal{A}}$ focus on \emph{when} to rely on each source, but $f_{def}$ assumes $Y$ to be always correct, whereas $f_{\mathcal{A}}$ uses estimated consistency as a criteria to determine which label to assume is correct. In contrast to choosing which label to rely on at different times, $f_{weak}$ and $f_{ens}$ blend both sources for all instances, which makes them much more sensitive to mistakes in either label, as shown by the variability of their results across settings. Finally, $f_{noise}$ relies on $D$ alone, missing out on the opportunity to learn from $Y$. These results show that in cases where expertise varies across instances, being judicious in determining which label to rely on can have a large payoff, and estimated expert consistency enables this goal. More broadly, these results illustrate that the proposed work constitutes an important, novel tool for managers interested in blending different sources of knowledge.
%


\begin{table}
\centering
\footnotesize
\addtolength{\tabcolsep}{-2pt}
\begin{tabular}[b]{lccccccc}
\midrule
\textit{\textbf{Scenarios}} & ${f}_Y$    & ${f}_D$  & ${f}_{def}$  & $f_{noise}$& $f_{ens}$ & $f_{weak}$ & ${f}_\mathcal{A}$ \\ \hline
Correct homo. & 0.795 (0.003) &  0.545 (0.024) &  0.516 (0.010) &  0.792 (0.007) &  0.767 (0.055) &  0.767 (0.004) &   \textbf{0.856} (0.008) \\
Corr. \& inc. homo.& 0.795 (0.003) &  0.437 (0.007) &  0.405 (0.005) &  0.670 (0.011) &  0.768 (0.044) &  0.771 (0.004) &   \textbf{0.860} (0.008) \\
Corr. \& inc. hetero.& 0.795 (0.003) &  0.551 (0.010) &  0.532 (0.011) &  0.812 (0.008) &  0.768 (0.044) &  0.766 (0.004) &   \textbf{0.860} (0.007)\\\bottomrule \end{tabular}
\captionof{table}{Methods' Area under the ROC Curve (AUC) with respect to $Y^c$ in different simulations of experts' decisions, corresponding to different assumptions of human expertise and errors. The proposed methodology outperforms all others across the different scenarios.}
\label{tab:sim_auc_err}
\end{table}

\subsubsection{Learning from biased experts}\label{subsub:mimic_biased_scenarios} We now explore what happens when shared human biases systematically underserve one demographic subgroup by making erroneous assessments. These experiments are motivated by the fact that, as we have stated throughout, the proposed approach assumes that consistency is indicative of correctness. 
However, the methodology is robust to shared biases that do not yield deterministic decisions. For example, if experts are more likely to make a certain decision for one group than for another, this bias will not be absorbed by the proposed methodology as long as the decision is not near-deterministic for either group. If, for instance, the parameter $\delta$ used to define the amalgamation set is chosen to be $\delta=0.95$, shared biases might be learned only if one subgroup exists for whom a decision is made 95\% of the time. The proposed methodology also is robust when biases yield deterministic decisions by one or a few experts but the biases are not widely shared. This section shows the performance of the method under various scenarios of bias. 

We assume the presence of a bias that increases the likelihood that women who require prioritization would be erroneously screened out. For the purpose of stress-testing the method, the settings we consider are probably more extreme than what may occur in practice. After considering multiple settings under which the method should in theory be robust, we consider a setting in which the method is bound to fail to emphasize the importance of determining that the method's assumptions are appropriate when choosing to apply it in practice. The following settings are assessed:
\begin{itemize}
    \item \emph{Deterministic bias, partially shared:} We assume that 50\% of experts exhibit a deterministic bias that leads them to screen out all women, regardless of whether they need prioritization. 
    \item \emph{Homogenous bias, fully shared:} We assume that all experts exhibit bias that leads them to screen out 80\% of women who should be prioritized. 
\item \emph{Non-random expert-to-patient assignment, near-deterministic bias:} We assume a non-random assignment under which one expert assesses 95\% of women and screens out all of them.
\item \emph{Deterministic bias, fully shared (method's assumptions violation):} Finally, we consider what happens when the method's assumptions are violated. In this setting, we assume all experts have a deterministic bias that leads them to screen out all women.
\end{itemize}

\begin{table}[t]
\centering
\footnotesize
\addtolength{\tabcolsep}{-2pt}
\begin{tabular}[b]{lccccccc}
\midrule
\textit{\textbf{Scenarios}}                     &${f}_Y$    & ${f}_D$  & ${f}_{def}$  & $f_{noise}$& $f_{ens}$ & $f_{weak}$ & ${f}_\mathcal{A}$ \\ \hline
Det. bias, part. sh.                & 0.794 (0.003) &  0.611 (0.013) &  0.502 (0.050) &  0.666 (0.006) &  0.741 (0.018) &  0.778 (0.004) &   \textbf{0.805} (0.008) \\
Hom. bias, fully sh. & 0.794 (0.003) &  0.612 (0.023) &  0.565 (0.018) &  0.690 (0.007) &  0.751 (0.021) &  0.777 (0.003) &   \textbf{0.806} (0.008) \\
Non-rand. assign.            &  0.794 (0.003) &  0.614 (0.008) &  0.535 (0.056) &  0.668 (0.006) &  0.741 (0.019) &  0.778 (0.003) &   \textbf{0.802} (0.008) \\ \hdashline
Det. bias shared                & \textbf{0.794} (0.003) &  0.544 (0.002) &  0.505 (0.030) &  0.678 (0.004) &  0.737 (0.015) &  0.773 (0.003) &   0.574 (0.042) \\\bottomrule \end{tabular}
\captionof{table}{Methods' Area under the ROC Curve (AUC) in different simulations of experts' decisions, corresponding to different assumptions about humans' shared bias. The proposed methodology is robust to biases, if these are not deterministic and shared across all experts.}
\label{tab:sim_auc_bias}
\end{table}
\begin{table}[t]
\centering
\footnotesize
\addtolength{\tabcolsep}{-2pt}
\begin{tabular}[b]{lccccccc}
\midrule
\textit{\textbf{Scenarios}}                     &${f}_Y$    & ${f}_D$  & ${f}_{def}$  & $f_{noise}$& $f_{ens}$ & $f_{weak}$ & ${f}_\mathcal{A}$ \\ \hline
Det. bias, part. sh. & \textbf{0.189} (0.002) &  0.120 (0.008) &  0.053 (0.033) &  0.145 (0.002) &  0.157 (0.003) &  0.185 (0.004) &   0.180 (0.003) \\
Hom. bias, fully sh. & \textbf{0.189} (0.002) &  0.109 (0.021) &  0.099 (0.011) &  0.148 (0.002) &  0.159 (0.004) &  0.183 (0.005) &   0.180 (0.004) \\
Non-rand. assign. & \textbf{0.189} (0.002) &  0.121 (0.005) &  0.075 (0.038) &  0.145 (0.002) &  0.158 (0.004) &  0.187 (0.004) &   0.179 (0.004) \\\hdashline
Det. bias shared    & \textbf{0.189} (0.002) &  0.078 (0.001) &  0.054 (0.019) &  0.147 (0.002) &  0.158 (0.003) &  0.181 (0.004) &   0.094 (0.028) \\\bottomrule \end{tabular}
\captionof{table}{Methods' True Negative Rate (TNR) for women, the group discriminated against in the simulations of experts' decisions, corresponding to different assumptions about humans' shared bias. The proposed methodology $f_\mathcal{A}$ presents a TNR similar to $f_Y$ even when $f_D$ presents significantly lower performance.}
\label{tab:sim_tnr_bias}
\end{table}

Table~\ref{tab:sim_auc_bias} shows results under these different scenarios, reporting AUC with respect to $Y^c$. Across all settings in which the method is expected to be robust, label amalgamation yields modest improvements or comparable performance to ${f}_Y$, and drastic improvements with respect to $f_D$. Once again, it is the only approach that is able to simultaneously outperform $f_Y$ and $f_D$. Meanwhile, all other approaches yield worse results than $f_Y$, indicating that they are unable to leverage expert knowledge, and instead bias in human labels lead them astray.

The proposed method should prevent errors, idiosyncrasies, and biases from being absorbed by the model (except those that violate the method's core assumption). Given that the bias considered in these simulations results in disproportionate false negative errors for women, its robustness to bias can be better studied through the true negative rate (TNR) for this demographic subgroup, which is reported in Table~\ref{tab:sim_tnr_bias}. The classification threshold is set to a 30\% screen-out rate, which corresponds to the screen-out rate in the real-world data.\footnote{Note that the relatively low TNR of ${f}_Y$ indicates that relying on observed outcomes alone leads to erroneous screen-outs.} Across all settings (except the failure mode), the TNR of the proposed approaches is comparable to that of ${f}_Y$, even when ${f}_D$ exhibits substantially worse TNR. 

 Importantly, if the method's core assumption is violated and if expert consistency is a result of bias that yields near-deterministic decisions and that is shared by all experts, the proposed approach is detrimental to overall performance, as shown in Table~\ref{tab:sim_auc_bias}. Moreover, the TNR for women, shown in Table~\ref{tab:sim_tnr_bias}, indicates that in this setting the proposed approach absorbs the experts' bias. This result highlights the importance of domain knowledge when deciding whether the proposed approach is suitable. In Appendix~\ref{app:subsec_bias} we propose a demographic parity assessment that can serve as a diagnostic to assess the potential risk of human consistency encoding bias, and we show how it applies to this scenario in \ref{app:bias_risk_sim}. Additionally, Appendix \ref{app:mimic_hyperparam} further explores the \textit{deterministic bias fully shared} scenario to assess the model's sensitivity to parameter choices in a biased setting.


\subsection{Child maltreatment hotline screening decisions}
\label{subsec:res_child}

Having studied the effectiveness of the approach in simulation settings, we now turn to real data from a high-stakes domain. 
In the United States, over 4 million calls concerning potential child neglect or abuse are received by child maltreatment hotlines every year~\citep{NCANDS}. Investigations are invasive and place considerable burden on families, and investigation resources are limited; thus, unnecessarily screening in cases is undesirable. Call workers who receive these calls are tasked with deciding which cases should be screened in for further investigation. We refer to the decision to screen in a call as $D$. 
The information communicated in a call can often be insufficient to establish whether a child is at risk, which has led many jurisdictions to grant call workers access to different databases, including criminal history and use of state public assistance programs. Efforts to increase the availability of historical information about children and adults involved in a call have been accompanied by an interest in the use of ML models for risk prediction, as parsing the large amounts of data available is difficult for humans.

One such system has already been implemented in Allegheny County, PA, USA, where child maltreatment hotlines record 15,000 calls each year. Using multi-system administrative data as covariates, a predictive model is trained to estimate the probability that an investigation will result in an out-of-home placement of the child. Out-of-home placement, which refers to the placement of a child in foster care, constitutes an observed outcome $Y$ that serves as an imperfect proxy for the construct of interest: whether a child is at risk of harm ($Y^c$). Consistent with the setting that we assume in this paper, the county trained and deployed a decision support model using historical data; administrative data $X$ of screened-in cases were used to predict observed outcome $Y$, which corresponds to a model ${f}_Y$.
In this section, we empirically study the risks of optimizing for an imperfect proxy and assess whether the proposed methodology can help mitigate this problem. We do so by training our own model ${f}_Y$, as well as the baseline and proposed models listed in Section~\ref{subsec:exp}, using data provided by Allegheny County.

\subsubsection{Construct gap}
\label{subsubsec:child_construct}
Call workers' decisions to screen in cases are sometimes mistaken, which has motivated the County to design and deploy an algorithmic decision support tool. In particular, the County is concerned about false negative errors, in which call workers may fail to screen in calls that warrant investigation. 
There is also a construct gap between the objective of the deployed algorithm and the construct of interest. The algorithm is trained to predict the probability of out-of-home placement, $Y$. Specifically, the observed label, $Y$, records whether out-of-home placement is observed in the 730 days following a call. However, the goal is to screen in cases where the child may be at risk ($Y^c$). Not all cases where there is a risk to the child rise to the level where a court would determine family separation (placement) is warranted, and a number of other interventions and forms of support are available to reduce the risk to the child without family separation. The county has chosen to prioritize this outcome because of its severity; in addition, the out-of-home placement decision is external to the Allegheny County Department of Human Services (DHS), and thus it provides a form of validation by entities external to the agency. However, this choice means that $Y$ may fail to identify cases in which the child is at risk and other available services and interventions may be effective for the child. This gap poses a potential risk in the context of algorithmic decision support because 
recent work has shown that deploying decision support tools that optimize for narrowly circumscribed outcomes may shift the criteria that people use for making decisions~\citep{green2021algorithmic}. In the child welfare context, such a drift could have important societal implications. In particular, it could shift the agency's attention away from interventions that reduce the risk to the child without requiring family separations.

\begin{figure}[ht]
\centering
\includegraphics[width=0.8\linewidth]{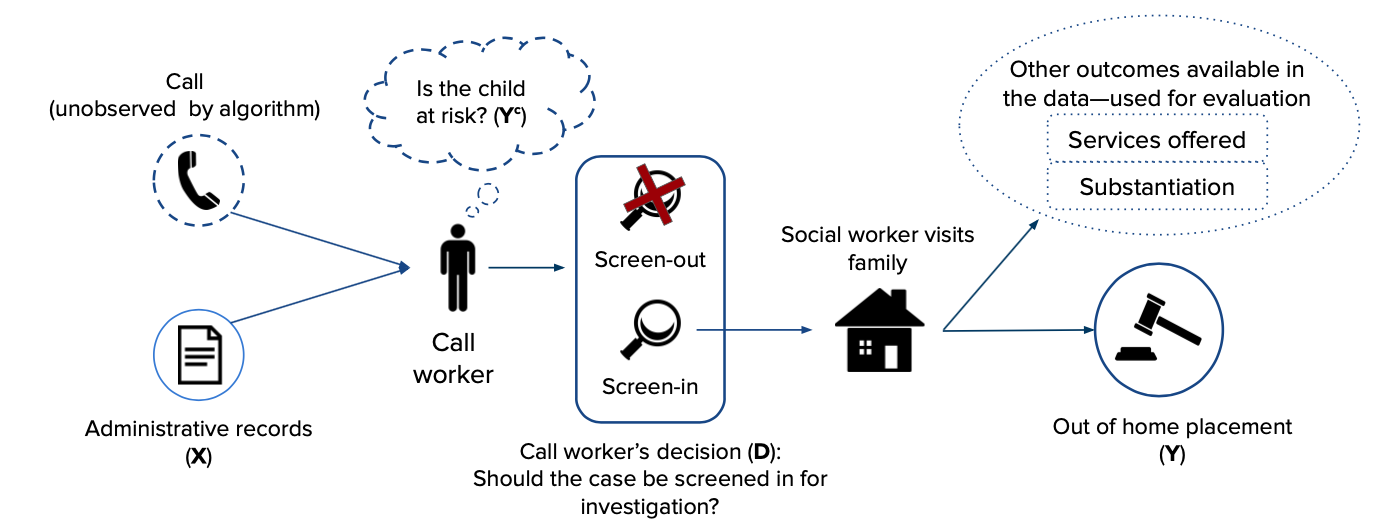}
\caption{Pipeline of child abuse hotline investigations.}
\label{fig:child_pipeline}
\end{figure}

The availability of other outcomes of interest that are recorded in the data but never observed by the model, either as covariates or as target labels, provides a unique opportunity to evaluate the proposed approach. The first of these outcomes is \emph{substantiation}, which encodes whether the social worker who visits the home deems the claims communicated in the call to be substantiated. The second outcome is \emph{services}, which records whether services are offered to the family as a result of the investigation and the family then receives voluntary or mandated services and monitoring aimed at reducing risk to the children. These outcomes, although also imperfect, shed light on other important dimensions that are highly relevant to the construct of interest to the agency: risk to the child. Figure~\ref{fig:child_pipeline} illustrates the pipeline of child abuse hotline investigations. 
In our experiments, we study whether existing models that seek to maximize performance but that have a narrow focus on out-of-home placement may actually incur a cost to performance with respect to substantiation and services. We then assess whether the proposed approach mitigates this problem. 

\subsubsection{Data}
\label{subsubsec:child_data}
The data used in our experiments correspond to $46,544$ calls received by Allegheny County between 2010 and 2014. This subset includes the first call associated with each child in this period of time, during which no risk assessment model was deployed. Call workers had access to the information communicated in the call, as well as direct access to the multi-system administrative data, including demographics, child welfare involvement, criminal history, and other information related to the children and adults associated with a call.  
These multi-administrative data were used by the county as covariates, $X$, to train deployed systems, and we also used them as covariates in our experiments. The data contain over 800 variables, including information regarding demographics, behavioral health, and past interactions with county prison and public welfare for all adults and children associated with a call. A detailed explanation of the Allegheny County Department of Human Services Data Warehouse and its use for predictive modeling can be found on the county's website~\citep{allegheny}.

\subsubsection{Predictive models and evaluation}
\label{subsubsec:child_models}
In this real-world setting we encounter the selective labels problem, as the outcomes of investigations are only available if an investigation is conducted in the first place. As a result, the system deployed by the county only uses screened-in cases ($D=1$) to train and evaluate the deployed model. We follow the same approach to train $f_Y$, and we use the entire dataset to estimate the high-consistency set $\mathcal{A}$ and to train all baselines, including $f_D$. 

As a result of the selective labels problem, for evaluation we are constrained by the fact that outcomes are only observed for screened-in cases. 
Thus, we do not use AUC, since this measure evaluates whether a model correctly identifies both negative and positive cases, but this cannot be appropriately assessed in this domain since we cannot determine whether screened-out cases were correctly classified by experts. 
Instead, we evaluate precision curves over the screened-in cases; this corresponds to showing precision when the classification threshold varies to include the top p\% highest scored screened-in cases. While not able to provide a complete oracle evaluation, this metric provides a path to assess the different models by comparing them on the basis of their relative prioritization of cases for which we observe labels. 
This metric assesses whether the highest ranked screened-in cases resulted in certain outcomes of interest. In other words, these precision curves allow us to answer the question: Among the cases that were screened-in, are those that resulted in a certain outcome of interest ranked higher than those that did not? For instance, among the 25\% of screened-in cases that score highest according to $f_Y$, what percentage resulted in out-of-home placement? What was this percentage for the other models? Note that this allows us to center our evaluation in the cases for which we know that the case warranted screen-in. 

In terms of parameter settings and training, our experimental setup is the same as described in Section~\ref{subsec:exp}, with $(\delta, \gamma_1, \gamma_2, \gamma_3) = (0.05, 4, 0.8, 0.002)$ (see sensitivity analysis and rationale for parameters' choice in Appendix~\ref{app:child_param}). We use neural networks following the training and evaluation procedure described in~\ref{subsec:exp}; results using a logistic regression are in Appendix~\ref{app:child_log}.

\subsubsection{Results}
\label{subsubsec:child_results}

\begin{figure}[t]
\centering
\includegraphics[width=0.7\linewidth]{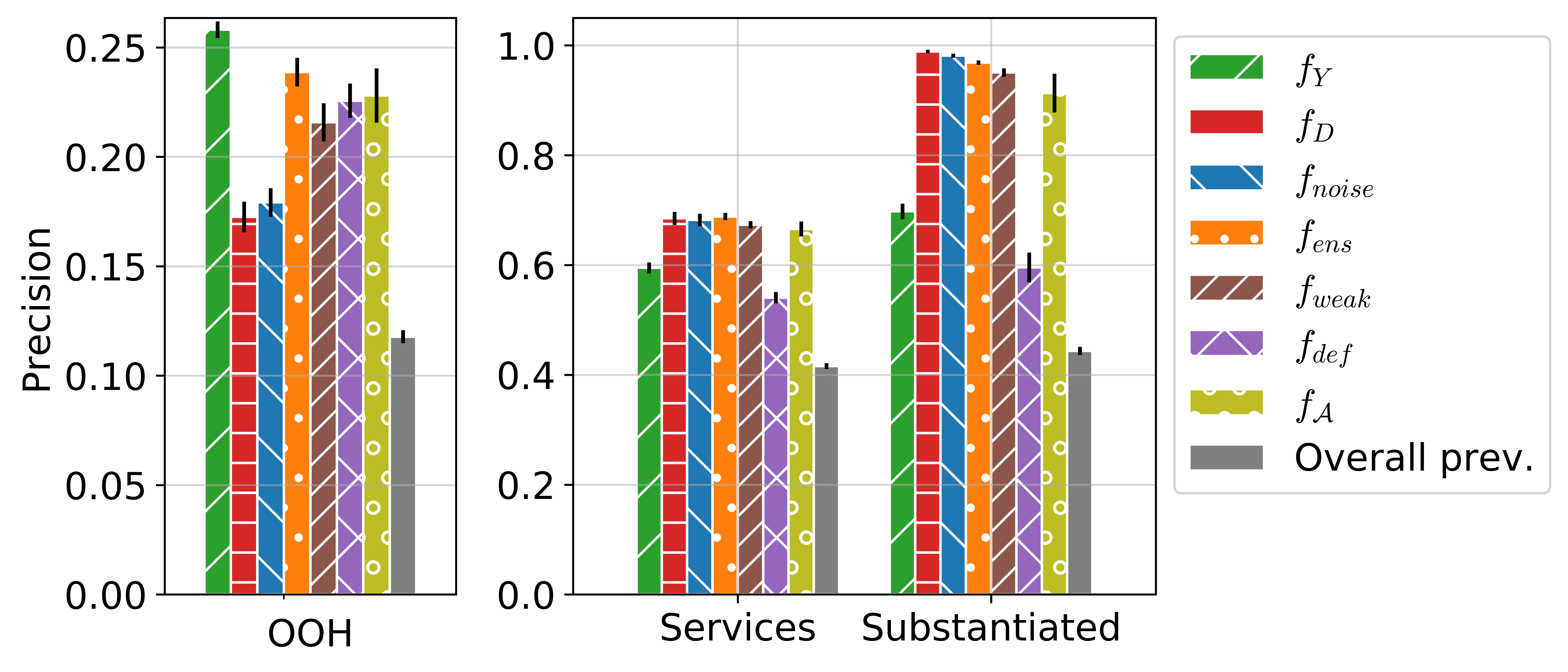}
\caption{Precision for top $25\%$ highest scored screened-in cases by model and outcomes. Error bars show the standard deviation over 10 runs of $75-25\%$ Monte Carlo cross-validation. Learning from $Y$ alone yields poor performance with respect to Services and Substantiated, while learning from expert decisions, $D$, alone yields poor performance with respect to OOH, even when applying a method robust to noise ($f_{noise}$). Combining both labels through different approaches, $f_{weak}$, $f_{ens}$ and $f_{\mathcal{A}}$ provides the ``best of both worlds".}
\label{fig:child}
\end{figure}

Figure~\ref{fig:child_topp_mlp} shows the precision curves of all models with respect to the observed outcome $Y$ (out-of-home placement) and to additional outcomes that are not observed by any of the models (services and substantiation). Some cut-offs have more practical relevance than others. Although the decision support tool does not provide binary recommendations, the screen-in rates have remained constant throughout the years~\citep{de2020case}. The county has resources to screen in approximately 40\% of the calls, and many calls are screened in solely on the basis of the information communicated in the call; therefore, it is reasonable to assume that cutoffs around 25\%-35\% have the most practical relevance. Figure~\ref{fig:child} shows precision at a 25\% cut-off.

\begin{figure}[t]
\centering
\includegraphics[width=0.9\linewidth]{Management-Science-template/fig/Welfare/nn_child.png}
\caption{Precision for top $p\%$ highest scored screened-in cases divided by outcome. Error bars show the standard deviation over 10 runs of $75-25\%$ Monte Carlo cross-validation. `Overall prev.' is the prevalence of each outcome. Learning from $Y$ alone yields poor performance with respect to Services and Substantiated, while learning from expert decisions, $D$, alone yields poor performance with respect to Out of Home placement (OOH), even when applying a method robust to noise ($f_{noise}$). Combining both labels through $f_{weak}$, $f_{ens}$ and $f_{\mathcal{A}}$ provides the ``best of both worlds". Which one is preferable depends on the specific threshold and sensitivity to the different outcomes. With respect to the priority outcome, OOH, $f_{\mathcal{A}}$ has the lowest decline in performance as thresholds increase.  
}
\label{fig:child_topp_mlp}
\end{figure}

\paragraph{Implications of optimizing for an imperfect proxy.} The model trained to predict out-of-home placement (${f}_Y$) does substantially better than the model trained to predict human decisions (${f}_D$) when evaluated with respect to out-of-home placement. This result indicates that call workers underperform at this task and that learning from observed outcomes has value. However, when comparing ${f}_Y$ and ${f}_D$ across other outcomes, we observe that a model trained on human decisions offers substantially higher performance in identifying cases that require services and those that are substantiated, while ${f}_Y$ fails to prioritize these cases. These findings indicate that there are cases in which risk is considered by call workers but not captured in the target labels used to train ${f}_Y$; as a result, this model exhibits poor performance for these cases. Thus, empirical evidence points to the need to narrow the construct gap in this domain because solely optimizing for out-of-home placement could risk a failure to support children and families who would benefit from services and other interventions.

\paragraph{Value of leveraging expert consistency.} The proposed label amalgamation, ${f}_\mathcal{A}$, successfully incorporates expert knowledge, and it considerably improves precision for cases that are substantiated and for which services are offered, compared to ${f}_Y$; meanwhile, it also retains performance with respect to out-of-home placement, substantially outperforming $f_D$ with respect to this outcome. This improvement can be seen in both Figure~\ref{fig:child_topp_mlp} and Figure~\ref{fig:child}. 

The ability of ${f}_\mathcal{A}$ to narrow the construct gap could have important practical implications. Consider the results for the top 25\% highest scoring screened-in cases, shown in Figure~\ref{fig:child}. In our data, this percentage corresponds to 1,565 cases recommended for screen-in. Among the cases selected for prioritization by ${f}_Y$ and ${f}_\mathcal{A}$, a similar number of cases resulted in out-of-home placement (an average of 403.9 and 356.9, respectively), which is markedly better than the number of cases prioritized by ${f}_D$ (270.1). Meanwhile, ${f}_\mathcal{A}$ prioritized 110.4 cases that were accepted for services and would not have been prioritized by ${f}_Y$ and 336.8 cases that were substantiated and would not have been prioritized by ${f}_Y$. In aggregate, among the 1,565 cases that each algorithm prioritizes, ${f}_Y$ flags 264.2 cases that have no associated outcome that may indicate risk, while ${f}_\mathcal{A}$ flags only 66.2 such cases. Thus, label amalgamation retains performance with respect to out-of-home placement, while narrowing the construct gap and prioritizing cases that may benefit from other types of intervention.

The performance of a learning to defer model, ${f}_{def}$, shows that its performance is comparable to ${f}_{\mathcal{A}}$ for out-of-home placement prediction but is significantly worse for substantiation and services. This result is not surprising, given that this method aims to improve performance with respect to $Y$. Thus, the results provide empirical evidence showing that using different criteria to combine observed outcomes and expert assessments can achieve significantly different goals.  

The performance of $f_{noise}$, which solely learns from $D$ but is robust to noise, shows poor performance with respect to out-of-home placement ($Y$), suggesting that the lower performance of $f_D$ with respect to this outcome is not simply due to noise, and emphasizing the value of jointly learning from $Y$ and $D$. 

In this context, $f_{ens}$ and $f_{weak}$, different alternatives to jointly learn from $Y$ and $D$, exhibit similar performance to the proposed methodology. This indicates that there may be multiple pathways to narrow the construct gap in this domain. The simulation experiments in Appendix~\ref{app:mimic_selective}, however, suggest that $f_{weak}$ may be particularly affected under selective labels, and thus may not be appropriate for this setting\footnote{Note that the underperformance of $f_{weak}$ under selective labels makes intuitive sense. For all screened-out instances, $f_{weak}$ effectively assumes that the call worker's decision was correct. Meanwhile, ${f}_{\mathcal{A}}$ only learns from cases where experts exhibit consistency, and $f_{ens}$ learns
two separate models that it then averages.}. From a managerial perspective, justification and implications of choosing one method over the other should be carefully considered. 

\paragraph{Normative grounding and procedural considerations} In this context, all three approaches that combine $D$ and $Y$ without assuming that $Y$ is a perfect outcome exhibit similar performance. However, their normative grounding and procedural considerations are significantly different, which is important for managers to consider. For instance, $f_{ens}$ averages the risk of out of home placement with the probability that humans would screen-in the call. $f_{weak}$ can be interpreted as doing something similar at the labeling stage. This does not align with what the county believes ought to be grounds for risk; in particular, risk of out-of-home placement alone is sufficient for the county to consider a case high risk. More importantly, in the presence of selective labels, $f_{weak}$ effectively assumes that the call worker's decision was correct \emph{every time}, whereas $f_{\mathcal{A}}$ takes a more conservative approach and only assumes call workers are correct when there is estimated consistency across them. This is especially important given that our evaluation focuses on the portion of screened-in cases, which could result in an overly optimistic depiction of the performance of $f_{weak}$, as suggested by the simulations studying the methods' robustness (or lack thereof) to the selective labels problem, shown in Appendix~\ref{app:mimic_selective}. Morevoer, one of the reasons why procedural considerations matter is because same overall performance does not mean that individual predictions are the same. While $f_{\mathcal{A}}$, $f_{ens}$ and $f_{weak}$ exhibit comparable performance with respect to all outcomes, the cases that they prioritize are significantly different. When considering the top 25\% highest scored screened-in cases by $f_{ens}$ and $f_{\mathcal{A}}$, there is only a 57.34\% overlap; when considering $f_{weak}$ and $f_{\mathcal{A}}$, this number is 51.83\%. The concept of predictive multiplicity~\citep{marx2020predictive} characterizes the fact that one dataset may accept multiple competing models with same overall performance but different grounds for classification. In such cases, how or why a prediction is reached has important procedural implications and may affect different individuals differently. In this domain, the proposed approach enables training models based on labels that consider a case as high risk if out-of home placement occurs \emph{or} if call-workers would consistently screen in the call, which is better aligned with the county's goals from a procedural perspective.

\paragraph{Model complexity} In some domains, including child welfare, less complex models are favored for deployment. Thus, relying on complex models, such as neural networks, can hamper the likelihood that a model will be adopted in practice. Furthermore, when using structured data, simpler models often can achieve comparable performance. Appendix~\ref{app:child_log} shows results when using logistic regression. The complexity of neural networks is important for ${f}_{def}$ and $f_{weak}$, while the other models, including the proposed $f_{\mathcal{A}}$, perform in a manner comparable to what was observed with neural networks. This result highlights the feasibility of applying the proposed methodology in contexts that favor limited model complexity.

\paragraph{Assessing risk of bias.} One concern in integrating estimated expert consistency in this setting is the need to avoid incorporating widespread stereotypes and biases held by call workers into the model. In particular, the presence of racial disparities would be worrying ~\citep{eubanks2018automating}. Table~\ref{tab:child} shows the racial breakdown when using a 25\% cut-off. Appendix \ref{app:child_log} shows the analogous result when using logistic regression. 
 These results provide a preliminary indication that performing label amalgamation in this setting does not exacerbate (nor does it mitigate) racial disparities in screen-in rates.

\begin{table}[ht]
\centering
\footnotesize
\addtolength{\tabcolsep}{-2pt}
\begin{tabular}[b]{lcccccccc}
\midrule
                     &${f}_Y$    & ${f}_D$  & ${f}_{def}$  & $f_{noise}$& $f_{ens}$ & $f_{weak}$ & ${f}_\mathcal{A}$ & Overall\\ 
Entire pop. & 48.85 (0.90) & 51.60 (0.64) & 50.92 (0.39) & 52.21 (0.61) & 51.71 (0.41) & 49.29 (2.22) & 50.32 (0.75) &  40.49 (0.43) \\
Screened-in & 52.68 (0.93)	& 50.67 (1.07) &	50.04 (0.55)&	51.30 (0.75)& 	53.12 (1.20) & 53.91 (2.24) & 50.77 (0.81) & 46.10 (0.53)   \\
\bottomrule
\end{tabular}
\captionof{table}{Percentage (std) of calls that involve at least one Black child. Results are shown when considering the top 25\% highest scored cases overall and the top 25\% highest scored screened-in cases. This result provides preliminary indication that the proposed label amalgamation does not introduce racial biases.}
\label{tab:child}
\end{table}
Overall, the results show that optimizing for out-of-home placement alone poses the risk of providing algorithmic recommendations that fail to identify cases where the child is at risk but where an out-of-home placement does not occur because other interventions are likely to be effective. In addition, this domain has provided us with a rare opportunity to access alternative outcomes that are relevant to the construct of interest but that are not reflected in the observed outcome chosen to train models that are currently in use. This access constitutes a unique opportunity to validate the proposed methodology. As our results show, learning from expert consistency using the proposed methodology can be a valuable pathway toward narrowing the construct gap.

\section{Discussion}
\label{sec:disc}
This paper tackles the construct gap that frequently exists between the construct of interest for a decision-making task and
the proxies available to train predictive models meant for decision support.  
The proposed approach narrows the construct gap by leveraging historical expert decisions stored in organizational information systems. 
In this work, we first propose a methodology to estimate consistency across experts on the basis of historical data that records a single decision per case. The proposed approach uses influence functions to enable consistency estimation in the presence of a non-random assignment of experts-to-instances, a common circumstance across application domains. Using the inferred expert consistency as a cornerstone, we then propose a label amalgamation approach to learn from experts in cases where experts exhibit consistency and to learn from observed outcomes elsewhere. 

Across all results, the proposed approach performs better or comparably to all baselines. Results on simulations show that the proposed methodology is successful across various scenarios of expert decision-making. Across all settings studied, except for the one that represents its failure mode, $f_{\mathcal{A}}$ outperforms all alternatives, and it is the only one that simultaneously outperforms $f_Y$ and $f_D$. This indicates that estimating and leveraging expert consistency allows the proposed approach to extract knowledge from observed human decisions, and not be led astray by human errors. More importantly, the simulations allow us to study a wide range of settings that may arise in practice, in which humans may make correlated mistakes or share biases. These results indicate that the conservative approach that $f_{\mathcal{A}}$ takes to learning from D pays off; by choosing to \emph{only} learn from experts when there is estimated consistency, $f_{\mathcal{A}}$ exhibits a more robust performance. While other baselines see significant drops in performance across some robustness tests, such as when humans share biases, $f_{\mathcal{A}}$ is only affected when humans have shared \emph{and} deterministic biases.

In the context of child maltreatment hotline screenings, a high-stakes setting in which algorithms are increasingly being deployed to support decisions across the US, ours is the first work to empirically characterize the construct gap. Our results indicate that some elements of risk are not wholly captured in the target label for which currently deployed models are optimized, although hotline call workers do recognize and prioritize these risks. 
We also empirically show that jointly learning from $Y$ and $D$ can improve precision for these cases, while retaining predictive power with respect to out-of-home placement. This emphasizes that simultaneously learning from observed outcomes and from expert assessments, which is rarely done in practice, provides valuable opportunities and should be considered more often. While multiple approaches perform comparably to the proposed approach in this domain, the superior performance and robustness of the proposed approach shown in the simulations indicates that $f_{\mathcal{A}}$ provides an overall superior performance. Moreover, its procedural considerations---only learning from expert consistency instead of learning from all expert decisions---, make this approach a more conservative and responsible way of incorporating information from expert assessments, as evidenced in the robustness tests. 

\subsection{Managerial Implications} Expert knowledge is a key asset for any organization, and an increased adoption of data storage and documentation practices has resulted in organizational information systems that store large amounts of historical expert decisions. These data are highly valuable but also very noisy because experts can be mistaken or biased and can disagree with one another. The difficulty in differentiating between signal and noise means that often these data remain unused. The proposed approach constitutes a path to unlock the value in such data. In particular, the proposed methodology broadens the options available to formulate the target objective of an algorithmic decision support system, providing an alternative that can narrow the construct gap. Crucially, this is achieved while relying solely on data that is typically available in organizational information systems, and does not necessitate the creation of expensive and time consuming expert panels. This provides a valuable mechanism to integrate expert knowledge in decision support systems, while being robust to their idiosyncrasies and errors. As shown in the empirical results, applying the proposed methodology to simultaneously learn from experts and from observed outcomes can help narrow the construct gap in comparison to commonly used target objectives. 

It is essential to note that the proposed methodology \emph{broadens} the options available when choosing a target objective, but it does not \emph{circumvent} the managerial task of making such a choice. Specifying a target objective always involves value judgements~\citep{coyle2020explaining}; integrating expert consistency or choosing to optimize for a single proxy are all design choices that come with their own limits. The proposed approach should only be chosen when managers deem expert consistency to be a desirable source of knowledge, as it is not suitable if experts are collectively and consistently optimizing for an undesirable objective. For example, if incentives are misaligned so that experts are consistently optimizing for an outcome but said outcome is aligned with their personal gain and not with the construct of interest, then the proposed methodology is not appropriate. Thus, managers are still responsible for assessing whether incentive structures are adequate. Both human experts and algorithms need managers~\citep{luca2016algorithms}, and the proposed methodology constitutes one more tool at managers' disposal to enable the design of better algorithmic decision support tools.  

\subsection{Future work}

The proposed work opens up several avenues for future research. In this section, we first remind readers of the central limitations of our work and outline how they could inform future work. We then discuss how the proposed methodology could be used to address a variety of other problems and emphasize that alternative solutions also may be available for the problems we have tackled.

\paragraph{Limitations.} The core assumption of our work is that experts' consistency is a valuable source of information. This assumption is reasonable in many applications, where experts' decisions are grounded in deep domain knowledge and scientific discoveries. However, learning from expert consistency may not always be desirable. 
The results in Section~\ref{subsec:semi} show that the approach is robust to many forms of shared biases, but they also provide empirical illustration of what may happen if bias shared among experts results in near-deterministic decisions. In Appendix~\ref{app:subsec_bias}, we have proposed a diagnostic approach that provides a partial assessment of the potential risk of bias. Future work should study the prevalence of near-deterministic expert biases, as well as alternative diagnostic approaches.

\paragraph{Other uses of labelers' influence estimation.} 
We have focused on one application of the proposed methodology to leverage expert consistency, however, we believe that the estimated influence can support more applications. 
Drawing connections to~\citet{geva2019}, for example, one could cluster experts based on similarity of influence, in order to derive insights about experts. Moreover, while we have focused on \emph{experts} [given that domain knowledge encoded in expert decisions makes consistency among them particularly valuable, we anticipate promising applications focused \emph{labelers'} influence more broadly. 
For example, there may be multiple uses in crowdsourcing, especially in contexts where labelers are not equally well positioned to assess different instances~\citep{davani2022dealing,fazelpour2022diversity}. Drawing connections to the role of influence functions in adversarial attacks~\citep{koh2017understanding}, the proposed approach could also be used to assess the vulnerability of a model to low-performing or adversarial labelers.

 \paragraph{Other approaches to estimate and leverage experts' consistency.} Throughout our work, we have made many methodological choices in deciding how to estimate consistency, as well as how to leverage it. We anticipate that multiple alternative avenues may be available to tackle the same problems. For instance, one could potentially develop approaches to estimate expert consistency through means other than influence functions. If enough instances per expert exist to train a classifier based on each of them, one could explore the use of boosting approaches to productively integrate them. Similarly, Bayesian approaches are a natural fit to integrate different sources of labels and could potentially be used both to estimate and leverage consistency. Future work that provides alternative ways of tackling the problems we outline could further improve organizations' ability to make productive use of expert decisions stored in their information systems. In particular, while we have focused on classification tasks, future work could expand the scope of the problem to leverage expert consistency in regression tasks.

\paragraph{Other implications for trust and deployment.} 
Our work is motivated by the need to narrow the construct gap when developing decision support tools. 
Other advantages to encoding estimated expert consistency in decision support tools are also apparent. First, different types of labels and informational sources may encode different biases, and thus jointly learning from them may reduce bias overall. Empirically and theoretically studying this potential outcome is a valuable direction for future work. In addition, future work may address the implications for experts' trust on algorithmic recommendations. Previous work has discussed algorithm aversion and the fact that humans easily lose trust in algorithmic tools after seeing them err~\citep{dietvorst2015algorithm}. Not all errors have the same effect; for example, when experts are very confident in their own assessment and see the algorithm err, such errors may be more detrimental to trust. Thus, future work could assess whether the proposed approach can mitigate algorithm aversion. Finally, incorporating experts' knowledge into algorithmic decision support tools may be especially useful when the experts whose knowledge is encoded into the algorithm are not the same as those using the algorithm. In particular, when algorithmic decision support tools are developed at highly specialized institutions and deployed widely, incorporating expert knowledge into the model may serve as a pathway to transfer knowledge from top experts to other facilities and organizations.

\newpage
\bibliographystyle{apalike}

\newpage

\begin{APPENDICES}
\label{appendix}




\section{Notation}
\label{app:notation}

\renewcommand{\arraystretch}{1.35}
\begin{table*}[ht]
  \centering
  \SingleSpacedXI
\small
\resizebox{\columnwidth}{!}{%
    \begin{tabular} { p{0.13\linewidth} p{0.87\linewidth}}
  \toprule
  \textbf{Notation}      & \textbf{Description } \\ \midrule
  $X$ &  Covariates observed. $X$ denotes the random variable; $\mathbf{X} \in  \mathbb{R}^{m \times n}$ denotes a matrix of training data; and $\mathbf{X}_v \in  \mathbb{R}^{k \times n}$ denotes a matrix of validation data. 
  \\ 
   $H$  & Expert who made decision observed in historical data.         \\ 
   $D$  & \textit{Decision}: Expert decision observed in historical data, binary.         \\ 
   $Y^c$  & \textit{Construct of interest}: Unobserved decision objective, binary.        \\ 
   $Y$  &  \textit{Observed outcome}: Observed outcome in historical data, binary. An imperfect proxy for $Y^c$.      \\ 
   $Y^{\mathcal{A}}$ & Amalgamated label, defined in Eq.~\eqref{eq:amalg}.\\

  \hdashline
  ${P}(\cdot)$ & Estimated probability that $\cdot$ will occur. 
  \\ 
  ${f}_D$ &  Model trained to predict expert decisions, ${f}_D={P}(D=1  \mid  X)$. 
  \\ 
  ${f}_Y$ &  Model trained to predict the observed outcome, ${f}_Y={P}(Y=1  \mid  X)$.  
  \\ 
  ${f}_{\mathcal{A}}$ &  Model trained to predict the amalgamated label, ${f}_{\mathcal{A}}={P}(Y^{\mathcal{A}}=1  \mid  X)$.   
  \\ 
  ${f}_{noise}$ & Model learning learning from noisy labels, defined in Section~\ref{subsec:exp}. \\
  ${f}_{def}$ & Deferral model, defined in Section~\ref{subsec:exp}. \\
  ${f}_{ens}$ & Ensemble model, defined in Section~\ref{subsec:exp}. \\
  ${f}_{weak}$ &  Weak supervision model, defined in Section~\ref{subsec:exp}. \\
  
  \hdashline
  $\delta$ &  Parameter denoting the level of consistency desired, $\delta \in [0.5,1]$.  \\
  $\mathcal{A}$ & Estimated consistency set. Variant $\mathcal{A}_{rand}$ defined in Eq.~\eqref{eq:A_def} under random expert-to-cases assignment and $\mathcal{A}_{non-rand}$ in Eq.~\eqref{eq:A_c} under non-random assignment. \\ 
  $\mathcal{A}_{0}$, $\mathcal{A}_{1}$ & Estimated $\delta$-level consistency set for $D=0$ and $D=1$, respectively. $\mathcal{A}=\mathcal{A}_{0}\cup \mathcal{A}_{1}$. \\ 
  \hdashline
  $\bm{w}$, $\bm{w}^h$ & 
  Weight vectors associated with instances in the training set. $\bm{w}=[1,...,1] \in \mathbb{R}^{m}$ is default weight; $\bm{w}^h\in \mathbb{R}^{m}$ is weight associated with upweighting expert $h$, defined in Eq.~\eqref{eq:w}. \\ 
  $\mathcal{I}_{D}(\bm{w}^h,\bm{x})$ & Normalized influence over ${f}_D(\bm{x})$ of upweighting expert $h$, defined in Eq.~\eqref{eq:inf_norm}.\\
  $\bm{s}(\bm{x})$ & Sorted absolute influence of experts on the prediction ${f}_D(\bm{x})$, defined in Eq.~\eqref{eq:s}
  ; $\bm{s}(\bm{x}) \in \mathbb{R}^q$, $s_j(\bm{x})$ denotes j$th$ entry of vector $\bm{s}(\bm{x})$.\\
  $m_1(\bm{x})$ & Metric for center of mass of influence, defined in Eq.~\eqref{eq:m1}. 
  \\
   $m_2(\bm{x})$ & Metric for opposing influence, defined in Eq.~\eqref{eq:m2}. 
   \\
    $m_3(\bm{x})$ & Metric for negligible maximum influence, defined in Eq.~\eqref{eq:m3}. 
    \\
  \bottomrule
  \end{tabular}
  }
    \caption{Notation used in this article.}
  \label{table:notation}
  \end{table*}

\section{Formalisation of expert consistency}
\label{app:consistency}

As stated in Section~\ref{sec:rand}, we say that experts exhibit $\delta$-level consistency for a given instance if there is a decision $d$ such that the probability that a randomly selected expert will make decision $d$ is greater or equal to $\delta$. This can be mathematically expressed as
\begin{equation}
    \frac{1}{q}\sum_h P(D=d|X=\bm{x}, H=h) \geq \delta.
\end{equation}

Note that the expression above accounts for the probability of uniformly sampling an expert, where the number of experts is $q$, and also accounts for the probability that the selected expert $h$ will make decision $d$ for instance $\bm{x}$. 
Under the assumption of randomised assignment of experts-to-instance, i.e. $X \ind H$, then $P(H=h \mid X=\bm{x}) = P(H=h) = \frac{1}{q}$. Therefore,
\begin{eqnarray}
\begin{array}{rl}
    \frac{1}{q}\sum_h P(D=d\mid X=\bm{x}, H=h) &= \sum_h P(D=d\mid X=\bm{x}, H=h)P(H=h) \\
    &=  \sum_h P(D=d\mid X=\bm{x}, H=h)P(H=h \mid X=\bm{x})\\
    & =  P(D=d\mid X=\bm{x}).
    \end{array}
\end{eqnarray}

Thus, when $X \ind H$, we can express $\delta$-level consistency for an instance $\bm{x}$ as
\begin{equation}
P(D=d \mid X=\bm{x}) \ge \delta.
 \end{equation}

\section{Proofs}

\subsection{Theorem~\ref{theorem:mis}}
\label{proof:mis}

\paragraph{Theorem:}
Given an estimated probability ${f}_D={P}(D=1|X)$ and a set $\mathcal{A}_{1}=\{\bm{x} \in \mathbf{X}  : {f}_D(\bm{x})\geq \delta \}$, the confidence interval of the true probability, $P(D=1|X)$, for data points in $\mathcal{A}_{1}$ can be estimated as:
 \begin{eqnarray*}
    \mathsf{CI}(P(D=1|X=\bm{x}, \bm{x}\in \mathcal{A}_{1})\geq \delta \hspace{0.01in} ;\hspace{0.01in} C) = \left(   \delta- z^{\ast}\frac{\sigma}{\sqrt{k}} ,  1 \right) , 
\end{eqnarray*}
where $CI(P; C)= (a,b)$ denotes that $(a,b)$ is the confidence interval of P at level C. Here, $\mathbf{X}_{v}\in \mathbb{R}^{k \times n}$ is a validation set drawn from the same distribution as $\mathbf{X}$; $k$ is the number of instances in this validation set; 
$\sigma$ is the standard deviation of the set $\{ d_i  \hspace{0.02in}  |  \hspace{0.02in}  \bm{x}_i\in \mathbf{X}_{v}, {f}_D(\bm{x}_i) \geq \delta\} $, which contains the expert decisions for instances in the set $\mathbf{X}_{v}$ that are estimated to lie in set $\mathcal{A}_{1}$; and $z^{\ast} = \Phi^{-1}(1-\frac{\alpha}{2})$ for $\alpha = \frac{1-C}{2}$ and $\Phi$ the standard normal distribution.

\proof{Proof:}
The confidence interval for the true probability $P$ can be bounded as shown in Equation~\eqref{eq:ci_raw}, which follows from the Central Limit Theorem, 
\begin{eqnarray}
    \mathsf{CI}(P(D=1|X=\bm{x}, \bm{x} \in \mathcal{A}_{1})\geq\delta \hspace{0.01in} ;\hspace{0.01in} C) = \left(  \mu - z^{\ast}\frac{\sigma}{\sqrt{k}} , \mu + z^{\ast}\frac{\sigma}{\sqrt{k}} \right) ,
    \label{eq:ci_raw}
\end{eqnarray}

 where $\mu$ corresponds to the mean of the set $\{ d_i  \hspace{0.02in}  |  \hspace{0.02in}  \bm{x}_i\in \mathbf{X}_{v}, {f}_D(\bm{x}_i) \geq \delta\} $.

 Note that since $\mathbf{X}$ and $\mathbf{X}_{v}$ are sampled from the same distribution, and ${f}_D$ is a calibrated probability distribution, then $\delta \leq \mu \leq 1 $. Hence,
 \begin{eqnarray}
    \mathsf{CI}(P(D=1| X=\bm{x}, \bm{x} \in \mathcal{A}_{1})\geq\delta \hspace{0.01in} ;\hspace{0.01in} C) = \left(   \delta- z^{\ast}\frac{\sigma}{\sqrt{k}} ,  1 \right).
\end{eqnarray}
\endproof

\subsection{Theorem~\ref{theorem:inf}:}
\label{proof:inf}
\paragraph{Theorem}
Assuming a doubly differentiable loss $\mathcal{L}$ and an invertible Hessian,
the normalized influence $\mathcal{I}_{D}(\bm{w}^h,\bm{x})$ of perturbing the training data $\mathbf{X}\in \mathbb{R}^{n\times m}$ in the direction $\bm{w}^h\in \mathbb{R}^{n}$ can be approximated as:
\begin{equation}
 \mathcal{I}_{D}(\bm{w}^h,\bm{x})  \approx  -\frac{1}{| \mathbf{X}_h|} \nabla_{\theta}{P}(d|\bm{x},\hat{\theta})^T [\nabla^2_{\theta}\mathcal{R}(\hat{\theta})]^{-1}
 \left[\sum_{i,\bm{x}_i \in  \mathbf{X}_h} \nabla_{\theta} \mathcal{L}(\bm{x}_i,d_i,\hat{\theta})\right],
\end{equation}

where $\nabla_{\theta}$ denotes a gradient with respect to $\theta$ and where $\mathcal{R}(\theta):= \sum_{i=1}^n \mathcal{L}(\bm{x}_i,d_i,\theta)$ is the empirical risk.

\proof{Proof:}
%
Recall that the normalized influence is defined as
\begin{equation*}
   \mathcal{I}_{D}(\bm{w}^h,\bm{x}) :=  \left. \frac{1}{| \mathbf{X}_h|} \frac{\partial {P}(d | \bm{x},\hat{\theta}_h)} { \partial \varepsilon}\right \rvert_{\varepsilon=0} 
\end{equation*}


Let $\frac{\partial \hat{\theta}_h}{\partial \varepsilon}$ denote the derivative of $\hat{\theta}_h$ with respect to $\varepsilon$. Using the chain rule, we can express the normalized influence as:
\begin{equation}
    \left. \frac{1}{| \mathbf{X}_h|} \nabla_{\theta}{P}(d|\bm{x},\hat{\theta}_h)^T \frac{\partial \hat{\theta}_h}{\partial \varepsilon} \right \rvert_{\varepsilon=0}\\
    \label{eq:infl_p}
\end{equation}

Recall that $\hat{\theta}_h:= \textrm{argmin}_{\theta \in \Theta} \sum_{i=1}^n w^h_i \mathcal{L}(\bm{x}_i,d_i,\theta)$ and thus, considering the definition of $\bm{w}^h$, it can be expressed as:
\begin{equation}
\hat{\theta}_h  = \textrm{argmin}_{\theta \in \Theta}\mathcal{R}(\theta) + \sum_{i,\bm{x}_i \in  \mathbf{X}_h}\varepsilon \mathcal{L}(\bm{x}_i,d_i,\theta)
\end{equation}
From the first order condition we obtain that:
\begin{equation}
0 = \nabla_{\theta}\mathcal{R}(\hat{\theta}_h) + \sum_{i, \bm{x}_i \in  \mathbf{X}_h} \varepsilon \nabla_{\theta} \mathcal{L}(\bm{x}_i,d_i,\hat{\theta}_h) 
\end{equation}

Note that as $\varepsilon \rightarrow 0$ it is the case that $\hat{\theta}_h  \rightarrow \hat{\theta}$. Therefore, the Taylor expansion centered around $\hat{\theta}$, defining $\Delta{\bm{w}^h} = \hat{\theta}_h -\hat{\theta}$, yields:
\begin{equation}
\begin{split}
0 = & \nabla_{\theta}\mathcal{R}(\hat{\theta})+\sum_{i, \bm{x}_i \in  \mathbf{X}_h} \varepsilon \nabla_{\theta} \mathcal{L}(\bm{x}_i,d_i,\hat{\theta}) \\ + & [\nabla^2_{\theta}\mathcal{R}(\hat{\theta})+\sum_{i, \bm{x}_i \in  \mathbf{X}_h} \varepsilon \nabla^2_{\theta} \mathcal{L}(\bm{x}_i,d_i,\hat{\theta}) ] \Delta{\bm{w}^h} + \textrm{h.o.t.} \end{split}
\end{equation}

Solving for $\Delta{\bm{w}^h}$ and making use of the fact that $\hat{\theta}$ minimizes $\mathcal{R}(\theta)$ and thus $\nabla_{\theta}\mathcal{R}(\hat{\theta}) = 0$, we obtain:
\begin{equation} 
\Delta{\bm{w}^h}  \approx  - [\nabla^2_{\theta}\mathcal{R}(\hat{\theta})+\sum_{i, \bm{x}_i \in  \mathbf{X}_h} \varepsilon \nabla^2_{\theta} \mathcal{L}(\bm{x}_i,d_i,\hat{\theta}) ]^{-1}
   [\sum_{i, \bm{x}_i \in  \mathbf{X}_h} \varepsilon \nabla_{\theta} \mathcal{L}(\bm{x}_i,d_i,\hat{\theta})].
\end{equation}

Let $A = \nabla^2_{\theta}\mathcal{R}(\hat{\theta})$, $B =\sum_{i, \bm{x}_i \in  \mathbf{X}_h} \nabla^2_{\theta} \mathcal{L}(\bm{x}_i,d_i,\hat{\theta})$, $C = \sum_{i, \bm{x}_i \in  \mathbf{X}_h} \nabla_{\theta} \mathcal{L}(\bm{x}_i,d_i,\hat{\theta})$. Then, 
\begin{eqnarray} 
\begin{array}{rl}
  \Delta{\bm{w}^h}  \approx & - [A+ \varepsilon B]^{-1} \varepsilon C \\
  = & - (I + \varepsilon A^{-1}B)^{-1} A^{-1} \varepsilon C \\
  = & - [\sum_{n=0}^{\infty}(-1)^n\varepsilon ^n (A^{-1}B)^n] A^{-1} \varepsilon C \\
  = & -(I-\varepsilon A^{-1}B)A^{-1}\varepsilon C + \textrm{h.o.t.} \\
  = & -\varepsilon A^{-1}C +  \textrm{h.o.t.}\\
  \Rightarrow  \Delta{\bm{w}^h}  \approx & - [\nabla^2_{\theta}\mathcal{R}(\hat{\theta})]^{-1}[\sum_{i, \bm{x}_i \in  \mathbf{X}_h} \nabla_{\theta} \mathcal{L}(\bm{x}_i,d_i,\hat{\theta})]\varepsilon 
  \end{array}
\end{eqnarray}

Since $\Delta{\bm{w}^h} = \hat{\theta}_h -\hat{\theta}$, and $\hat{\theta}$ does not depend on $\varepsilon$, we get that
\begin{equation} 
\frac{\partial \hat{\theta}_h}{\partial \varepsilon}  =  \frac{\partial \Delta{\bm{w}^h}} {\partial \varepsilon} \approx  - [\nabla^2_{\theta}\mathcal{R}(\hat{\theta})]^{-1}[\sum_{i, \bm{x}_i \in  \mathbf{X}_h} \nabla_{\theta} \mathcal{L}(\bm{x}_i,d_i,\hat{\theta})].
\end{equation}

Replacing this in Equation~\eqref{eq:infl_p} yields:
\begin{equation}
 \mathcal{I}_{D}(\bm{w}^h,\bm{x})  \approx  - \frac{1}{| \mathbf{X}_h|} \nabla_{\theta}{P}(d|\bm{x},\hat{\theta})^T [\nabla^2_{\theta}\mathcal{R}(\hat{\theta})]^{-1}
 [\sum_{i, \bm{x}_i \in  \mathbf{X}_h} \nabla_{\theta} \mathcal{L}(\bm{x}_i,d_i,\hat{\theta})].
 \label{eq:infl2}
\end{equation}

\endproof

\section{Algorithmic Summary}
\label{app:algo_summary}

The proposed methodology consists of a series of steps, illustrated in Figure~\ref{fig:summary}. Algorithm~\ref{alg:steps} summarizes the steps involved. Assume we have a training dataset composed of covariates $\mathbf{X}\in \mathbb{R}^{m \times n}$, associated experts $H\in \mathbb{R}^{m \times 1}$, expert decisions $D\in \mathbb{R}^{m \times 1}$, observed outcomes $Y\in \mathbb{R}^{m \times 1}$. As a first step, we use cross-validation to estimate the influence of each expert over each instance. To do this, we train a model $f_D$, apply it to predict expert decisions, and estimate the influence of each expert over each prediction using Eq.~\eqref{eq:influence}. Once we have done this over all folds of the cross-validation, we have a matrix of estimated influence of dimensions ${m \times q}$. The next step is to estimate the high-consistency set $\mathcal{A}$, which we do by applying Eq.~\eqref{eq:A_c}. Once the high-consistency set is identified, we can define the amalgamated label $Y^\mathcal{A}$, which is equal to $D$ for instances in the set $\mathcal{A}$, and equal to $Y$ for all other instances, as defined in Eq.\eqref{eq:amalg}. Finally, we train the model $f_\mathcal{A}$ to predict $Y^\mathcal{A}$ from $X$. This model can then be evaluated in a separate test set and applied. 

\begin{algorithm}[ht!]
\caption{Leveraging Expert Consistency.}
\KwData{Training data: Covariates $\bm{X}\in \mathbb{R}^{m \times n}$, Associated Experts $H\in \mathbb{R}^{m \times 1}$, Expert Decisions $D\in \mathbb{R}^{m \times 1}$, Observed Outcomes $Y\in \mathbb{R}^{m \times 1}$.}
\KwResult{Estimator $f_\mathcal{A}$ for the construct of interest $Y^c$.}
\For{$(train, test) \in CrossVal(X)$}{
    $f_D \gets$ Model($D[train] \sim X[train]$); Model the observed expert decisions\;
    $f_D(X[test]) \gets$ Estimate expert decisions\;
    \For{$h \in \{1,...,q\}$}{
    $\mathcal{I}_{f_D}[w^h,test] \gets$ Estimate each experts' influence on the predicted decision (Eq.~\eqref{eq:influence})\;
    }
}
$\mathcal{A} \gets$ Identify instances with high consistency between experts using the previous quantities $\mathcal{I}_{D}$ and $f_D(\bm{X})$ (Eq.~\eqref{eq:A_c}) \;
$Y^\mathcal{A} \gets $ Amalgamate observed decisions when experts are consistent, and observed outcomes when they are not (Eq.\eqref{eq:amalg})\;
$f_\mathcal{A} \gets$ Model($Y_\mathcal{A} \sim X$); Model the amalgamated labels given the covariates\;
\label{alg:steps}
\end{algorithm}

As practical considerations, we note a few things. The models used for $f_D$ and $f_\mathcal{A}$ need not be of the same family, and the constraints for $f_D$ discussed in Section~\ref{subsec:inf}, such as the loss function being doubly differentiable, do not apply to $f_\mathcal{A}$. For instance, $f_D$ could be a neural network while $f_\mathcal{A}$ could be a random forest. Additionally, one can evaluate how well different types of models estimate consistency. As discussed in Section~\ref{sec:rand}, modeling choices may affect our ability to identify instances that belong to the $\delta$-level consistency set. Thus, testing different models and choosing the one that has the best predictive power for $D$ is desirable. Similarly, when considering how to model $f_\mathcal{A}$, it is a good idea to test a few models and evaluate their performance with respect to $Y^\mathcal{A}$. This is not different from what one would do when building any machine learning system and assessing what type of model fits the data best. 

\section{Asymmetric Amalgamation}
\label{app:asymmetric}

The amalgamation can be done on the entire set $\mathcal{A}$, or asymmetrically, only amalgamating one of two possible decisions. Depending on the context and the conceptual relationship between $Y$ and $Y^c$, it may be desirable to do asymmetric amalgamation, meaning only amalgamate decisions $D=0$ or $D=1$. For example, there may be cases where $Y^c=1$ whenever $Y=1$, and the construct gap only affects cases where $Y=0$ but $Y^c=1$. In such instances, one may choose to only amalgamate instances in which it is inferred that experts consistently make the decision $D=1$, which could improve recall for the subset of cases for which $Y^c=1, Y=0$. Such asymmetric amalgamation can be achieved by only amalgamating instances in the set $\mathcal{A}_{1}$. Analogously, one could choose to only amalgamate the set $\mathcal{A}_{0}$.

\section{Diagnosis of bias risk} 
\label{app:subsec_bias}

Although no diagnostic approach can conclusively determine whether estimated expert consistency reflects human biases, and domain knowledge is crucial to guide the responsible use of the proposed methodology, demographic parity assessment can be used as a diagnostic approach to determine \emph{potential} bias. Demographic parity is a measure of algorithmic fairness that considers the difference in rates of a given prediction across groups~\citep{chouldechova2017fair}. Its main operational advantage lies in the fact that it does not require access to a ``ground truth" label. However, this flexibility also constitutes its main disadvantage: differences in rates are not conclusively indicative of disparities in error rates because base rates may differ across groups. Let $\mathcal{G}$ denote a sensitive group (e.g., a racial minority), and let $\tau$ denote a classification threshold. The demographic parity gap $\textrm{Gap}_{dp}$ for predictive model ${f}~:~\mathbb{R}^{1 \times m}\rightarrow~[0,1]$ is given by
 \begin{eqnarray}
\textrm{Gap}_{dp}({f}(\cdot),\tau, \mathcal{G})=\frac{1}{|\{i \in \mathcal{G}\}|}\sum\limits_{i\in \mathcal{G}}\mathbbm{1}({f}(\bm{x}_i)>\tau)-\frac{1}{|\{i \notin \mathcal{G}\}|}\sum\limits_{i\notin \mathcal{G}}\mathbbm{1}({f}(\bm{x}_i)>\tau).
\label{eq:dem_gap}
 \end{eqnarray}

Assessing how the demographic parity gap is affected by label amalgamation can be useful. Let $\tau_Y$ and $\tau_\mathcal{A}$ denote classification thresholds for ${f}_Y$ and ${f}_{\mathcal{A}}$, respectively, such that these thresholds yield the same overall selection rate---that is, they predict a fixed percentage of the data as positive. For example, in the child welfare setting, one may want to compare classifiers when recommending screen-in for the top 25\% of the highest scoring cases. The shift in the demographic parity gap is given by
 \begin{eqnarray}
\Delta  \textrm{Gap}_{dp}= \textrm{Gap}_{dp}({f}_{\mathcal{A}}(\cdot),\tau_{\mathcal{A}}, \mathcal{G})-\textrm{Gap}_{dp}({f}_Y(\cdot),\tau_Y, \mathcal{G}).
\label{eq:dem_gap_diff}
 \end{eqnarray}

This measure can serve as an assessment of \emph{potential} bias, and swapping ${f}_Y$ for $f_D$ yields the assessment relative to human decisions. A change in the demographic parity gap is not necessarily indicative of bias. For example, in the context of predicting patients in need of high-cost care management programs, researchers found that the observed outcome that was chosen as a proxy target for a predictive model disadvantaged Black patients~\citep{obermeyer2019dissecting}. In such a case, if integrating expert consistency allows us to better approximate the construct of interest, then the demographic parity gap should change. However, changes that align with societal biases and that cannot be explained by domain knowledge may be indicative of a potential risk of bias. For example, in the child welfare domain, we consider whether leveraging inferred expert consistency significantly changes the racial and socioeconomic distribution of screened-in cases, thus reifying potential biases in the child welfare system~\citep{eubanks2018automating}. We also apply this diagnostic to the simulations, and show results in Appendix~\ref{app:bias_risk_sim}.  

\section{Neural networks training protocol}
\label{app:training_det}

For all models, we use multilayer perceptrons with ReLU activation functions. We select the number of hidden layers using cross-validation over the grid $[1,2,3]$, with all layers containing 50 neurons. Models were trained using binary cross-entropy for a maximum of 1,000 epochs, with batch size 100, an Adam optimizer with learning rate 0.001, and a dropout of 0.5. We also used an early stopping criterion: an increased loss across three iterations on a holdout set, defined as 15\% of the training split. (The holdout set is not used in training.).
When applying the proposed methodology to estimate expert consistency, we ensure the existence of the inverse of the Hessian by adding the smallest L$_1$ regularization necessary, iteratively exploring values in $[10^{-3}, 10^{-2}, 10^{-1}, 10^{0}, 10^{1}, ...]$. To prevent information leakage between the points used for model training and consistency estimates, we estimate the influence for each point within an internal 5-fold cross-validation framework. To ensure calibration, we use Platt scaling~\citep{platt1999probabilistic}. We use Monte Carlo cross-validation, with splits of 75\% train--25\% test, and we report normal confidence bounds at the 95\% level, estimated over 10 iterations. In all splits, we ensure that all experts have representation across all folds, and in the child welfare scenario, we also ensure that if a call involves multiple children, they are all in the same fold.

\section{Simulation experiments}

\subsection{Simulations of different scenarios}
\label{app:mimic_generation}

This section describes the simulations presented in Section~\ref{sec:res}. Using the MIMIC-ED data, we use vital signs measured at admission to the emergency room as covariates $\bm{X}$. To ensure that we have full control over the relationship between the covariates and outcomes, and to model different scenarios of human knowledge, errors, and biases, we simulate labels $Y$, $Y^{omitted}$ and $D$, as well as the assignment $H$, using tree-based models. In particular, let $\tilde{Y}^{omitted} = (\textrm{age}>65\textrm{ }or\textrm{ pain}\geq 7$) as observed in the data, and $\tilde{Y}$ hospitalization observed in the data. As described in the main body of the paper, all labels in our experiments are simulated. Specifically, we train a a decision tree to predict $\tilde{Y}^{omitted}$, and adopt its binary predictions as the simulated label $Y^{omitted}$. Then, to simulate anti-correlation between $Y$ and $Y^{omitted}$, we train a decision tree to predict $\neg \tilde{Y}$, where $\neg$ denotes negation, and adopt its binary predictions as the simulated label $Y$, as described in Algorithm~\ref{alg:cho_pc}. This anti-correlation simulation complements the results from real data, in which observed and omitted outcomes are positively correlated. We treat $Y$ as the simulated label of hospitalization and $Y^{omitted}$ as the simulated label of omitted risk (i.e., cases that require prioritization but do not result in hospitalization), such that $Y^c = Y \lor Y^{omitted}$. 
None of the covariates used as targets to simulate $Y$ and $Y^{omitted}$ are included in the covariates $\bm{X}$ made available to the predictive algorithms in our experiment.

We simulate the expert decisions $D$ using the following three scenarios:
\begin{enumerate}
    \item \emph{Correct homogenous beliefs and randomness:} Experts are bad at assessing patients who will be hospitalized ($Y$) but who do not present markers of omitted risk ($Y^{omitted}$), making random decisions in these cases. Experts are highly accurate (with respect to $Y^c$) for all other cases. 
    To simulate this, we train a decision tree to predict $Y^c$, and we sample a random decision for patients in leaves with high prevalence of $Y$ and low prevalence of $Y \lor Y^{omitted}$, denoted $\mathcal{L}$. For all other patients, we let $D$ be the prediction of the tree trained to predict $Y^c$. Details are included in Algorithm~\ref{alg:cho_pc}.
    \item \emph{Correct and incorrect homogenous beliefs:} In this setting, experts present a similar behavior. However, they incorrectly assess patients who will be hospitalized ($Y$) but who do not present markers of omitted risk ($Y^{omitted}$), making the wrong decision with regard to hospitalisation in 75\% of these cases. 
    This algorithm relies on the same labels generation than Algorithm~\ref{alg:cho_pc}. However, instead of drawing a random decision for all patients in $\mathcal{L}$, the negation of the observed outcome $Y$ is assigned to 75\% of them. For all other patients, we let $D$ be the prediction of the tree trained to predict $Y^c$. Details are included in Algorithm~\ref{alg:ciho_pc}.
    
    \item \emph{Correct and incorrect heterogeneous beliefs:} This scenario presents different error rates for each expert instead of the fixed 75\% error rate. For each expert, their error rate is uniformly drawn from an interval $[\rho, \rho+0.3]$. We consider three variants of this scenario, with $\rho=0.3$, $\rho=0.5$ and $\rho=0.7$. Algorithm~\ref{alg:cihe_pc} describes the associated simulation. The algorithm iterates over each expert. For each expert, it identifies patients in $\mathcal{L}$ who have been assigned to that expert. Subsequently, for these patients, the negation of their hospitalization outcome $Y$ is used as the expert decision with probability equal to the error rate specific to that expert. For all other patients, we let $D$ be the prediction of the tree trained to predict $Y^c$.  
    
\end{enumerate}

\begin{algorithm}[ht!]
\caption{Correct homogenous beliefs and randomness.}\label{alg:cho_pc}
\KwData{Vital signs measured at triage $\bm{X}$}
\KwResult{Simulation of labels $Y$ and $Y^{omitted}$, expert decisions $D$ and assignments $H$}
$Y \gets$ Tree$_1$($\neg\Tilde{Y} \sim X$, depth = 15)\;
$Y^{omitted} \gets$ Tree$_2$($\Tilde{Y}^{omitted} \sim X$, depth = 15)\;
$Y^c \gets  Y \lor Y^{omitted}$\;
$H \gets$ Random (uniform) draws with replacement in $[\![1, q]\!]$\;
$D \gets$ Tree$_3$($Y^c \sim X$, depth = 10)\;
$\mathcal{L} \gets \{$Instances in leaves of Tree$_3$ that have prevalence of $Y$ $\geq$ 70\% and $Y \lor Y^{omitted}$ $\leq$ 30\%$\}$\;
\For{$\bm{x}_i \in \mathcal{L}$}{
    $D_i \gets \mathcal{B}ernoulli(0.5)$; Random draw of the human decision for instances in $\mathcal{L}$\;
}
\end{algorithm}

\begin{algorithm}[ht!]
\caption{Correct and incorrect homogenous beliefs.}\label{alg:ciho_pc}
\KwData{Vital signs measured at triage $\bm{X}$}
\KwResult{Simulation of labels $Y$ and $Y^{omitted}$, expert decisions $D$ and assignments $H$}
Obtain $Y^c$, $Y$, $Y^{omitted}$, $D$, $\mathcal{L}$, $H$ by executing lines 1 to 6 from Algorithm \ref{alg:cho_pc}\;
$\mathcal{L}_N \gets$ Randomly draw $75\%$ of instances in $\mathcal{L}$\;
\For{$\bm{x}_i \in \mathcal{L}_N$}{
    $D_i \gets \neg Y_{i}$\;
}
\end{algorithm}

\begin{algorithm}[ht!]
\caption{Correct and incorrect heterogeneous beliefs.}\label{alg:cihe_pc}
\KwData{Vital signs measured at triage $\bm{X}$ and base error rate $\rho$}
\KwResult{Simulation of labels $Y$ and $Y^{omitted}$, expert decisions $D$ and assignments $H$}
Obtain $Y^c$, $Y$, $Y^{omitted}$, $D$, $\mathcal{L}$, $H$ by executing lines 1 to 6 from Algorithm \ref{alg:cho_pc}\;

\For{$h \in [\![1, q]\!]$}{
    $e_h \gets \mathcal{U}([\rho, \rho + 0.3])$; Draw random error rate for expert $h$\;
    $\mathcal{L}_N \gets$ Randomly draw $e_h$ of instances in set $\{\bm{x}_i \in \mathcal{L}: H_i = h\}$\;
    \For{$\bm{x}_i \in \mathcal{L}_N$}{
        $D_i \gets \neg Y_{i}$\;
    }
}
\end{algorithm}

\newpage
For the scenarios with embedded biases, introduced in Section~\ref{subsub:mimic_biased_scenarios}, expert decisions depend on the group membership, $M$, with $M_i = 1$ corresponding to patient $i$ being a woman. For these experiments, all models trained on the simulated data take as inputs $\bm{X}$ and $M$, in order to capture potential relationships between group membership and outcome. We consider the following scenarios:

\begin{enumerate}
    \item \emph{Deterministic bias, partially shared}: 50\% of experts exhibit a deterministic bias that leads them to screen out all women ($M$). To simulate this, Algorithm~\ref{alg:dep_pc} uses the labels simulated in Algorithm~\ref{alg:cho_pc}. Then, it selects female patients associated with half of the experts, and assigns them the screen out decision.
    \item \emph{Homogenous bias, fully shared}: All experts exhibit bias that leads them to screen out 80\% of women. This scenario differs from the previous one as all experts --- instead of half of them --- present the same bias. To simulate this, Algorithm~\ref{alg:hof_pc} initially uses the labels simulated in Algorithm~\ref{alg:cho_pc}, and then assigns a screen out decision to 80\% of women, regardless of their actual outcomes of their associated expert assignment.
    \item \emph{Non-random expert-to-patient assignment, near-deterministic bias}: This scenario consists of an imbalance in cases' assignment. One expert assesses 95\% women, and decides to deprioritise all of them. Algorithm~\ref{alg:nrnd_pc} differs from the two previous scenarios as it alters the generation of $H$ with 95\% of women assigned to one expert. A screen-out decision is then assigned to these patients. The remaining experts are randomly assigned to the remaining patients, and their decisions are generated as described in Algorithm~\ref{alg:cho_pc}.
    \item \emph{Deterministic bias, fully shared (method's assumptions violation):} In this setting, we assume all experts have a deterministic bias that leads them to screen out all women. Algorithm~\ref{alg:des_pc} assigns to all women a screen-out decision regardless of their associated expert assignment. For men, decisions are assigned as described in Algorithm~\ref{alg:cho_pc}.
\end{enumerate}

\begin{algorithm}[ht!]
\caption{Deterministic bias, partially shared.}\label{alg:dep_pc}
\KwData{Vital signs measured at triage $\bm{X}$, Minority membership $M$}
\KwResult{Simulation of labels $Y$ and $Y^{omitted}$, expert decisions $D$ and assignments $H$}
Obtain $Y^c$, $Y$, $Y^{omitted}$, $H$ and $D$ by executing lines 1 to 5 from Scenario \ref{alg:cho_pc}\;
$\mathcal{L}_N \gets \{\bm{x}_i \in \bm{X}: H_i < \frac{q}{2} \land M_i = 1\}$; Select all women assigned to the first half of the experts\;
\For{$\bm{x}_i \in \mathcal{L}_N$}{
    $D_i \gets 0$\; 
}
\end{algorithm}

\begin{algorithm}[ht!]
\caption{Homogenous bias, fully shared.}\label{alg:hof_pc}
\KwData{Vital signs measured at triage $\bm{X}$, Minority membership $M$}
\KwResult{Simulation of labels $Y$ and $Y^{omitted}$, expert decisions $D$ and assignments $H$}
Obtain $Y^c$, $Y$, $Y^{omitted}$, $H$ and $D$ by executing lines 1 to 5 from Scenario \ref{alg:cho_pc}\;
$\mathcal{L}_N \gets $ Draw $80\%$ of instances in protected group $\{i, M_i = 1\}$\;
\For{$\bm{x}_i \in \mathcal{L}_N$}{
    $D_i \gets 0$\; 
}
\end{algorithm}

\begin{algorithm}[ht!]
\caption{Non-random expert-to-patient assignment, near-deterministic bias.}\label{alg:nrnd_pc}
\KwData{Vital signs measured at triage $\bm{X}$, Minority membership $M$}
\KwResult{Simulation of labels $Y$ and $Y^{omitted}$, expert decisions $D$ and assignments $H$}
Obtain $Y^c$, $Y$, $Y^{omitted}$, $H$ and $D$ by executing lines 1 to 5 from Scenario \ref{alg:cho_pc}\;
$\mathcal{L}_N \gets $ Draw $95\%$ of instances in protected group $\{i, M_i = 1\}$\;
\For{$\bm{x}_i \in \mathcal{L}_N$}{
    $H_i \gets 0$; Assign to expert $0$\;
    $D_i \gets 0$\;
}

\end{algorithm}

\begin{algorithm}[ht!]
\caption{Deterministic bias, fully shared.}\label{alg:des_pc}
\KwData{Vital signs measured at triage $\bm{X}$, Minority membership $M$}
\KwResult{Simulation of labels $Y$ and $Y^{omitted}$, expert decisions $D$ and assignments $H$}
Obtain $Y^c$, $Y$, $Y^{omitted}$, $H$ and $D$ by executing lines 1 to 5 from Scenario \ref{alg:cho_pc}\;
\For{$\bm{x}_i \in \{i, M_i = 1\}$}{
    $D_i \gets 0$\;
}
\end{algorithm}

\subsection{Further simulations of correct and incorrect erroneous beliefs}
\label{app:corr_incorr}

In the simulations, the setting \emph{Correct and incorrect heterogeneous beliefs} assumes that not all experts are equally likely to make a correct decision, and instead, their performance varies. In particular, we assume that when assessing patients who will be hospitalized and who do not present markers of the omitted risk, the probability of error with respect to hospitalization in these cases is $p_h$, for $p_h \sim \mathcal{U}(a,b)$. We consider three variations of this setting:

  \begin{itemize}
        \item[--] $p_h \sim \mathcal{U}(0.3,0.6)$. Some experts have above-random performance for this group, and some have worse-than-random performance. Above-random performance is more common.
        \item[--] $p_h \sim \mathcal{U}(0.5,0.8)$. All experts have worse-than-random performance for this group. The worst performing expert still correctly identifies at least 20\% of patients among this group who need prioritization. 
        \item[--] $p_h \sim \mathcal{U}(0.7,1.0)$. All experts systematically misdiagnose patients in this group. Some experts may erroneously screen out \emph{all} patients they assess from this pool. 
    \end{itemize}

All variations show consistent results, presented in Table~\ref{tab:sim_auc_err_app}.
    
\begin{table}
\centering
\footnotesize
\addtolength{\tabcolsep}{-2pt}
\begin{tabular}[b]{lcccccccc}
\midrule
\textit{\textbf{Scenarios}}                     &${f}_Y$    & ${f}_D$  & ${f}_{def}$  & $f_{noise}$& $f_{ens}$ & $f_{weak}$ & ${f}_\mathcal{A}$ \\ \hline
$\mathcal{U}(0.7 - 1)$  & 0.795 (0.003) &  0.426 (0.006) &  0.403 (0.004) &  0.624 (0.009) &  0.768 (0.044) &  0.770 (0.003) &   \textbf{0.860} (0.006) \\
$\mathcal{U}(0.5 - 0.8)$  & 0.795 (0.003) &  0.480 (0.016) &  0.416 (0.006) &  0.737 (0.009) &  0.768 (0.044) &  0.770 (0.003) &   \textbf{0.860} (0.009) \\
$\mathcal{U}(0.3 - 0.6)$  &  0.795 (0.003) &  0.551 (0.010) &  0.532 (0.011) &  0.812 (0.008) &  0.768 (0.044) &  0.766 (0.004) &   \textbf{0.860} (0.007) \\ \bottomrule \end{tabular}
\captionof{table}{Methods' Area under the ROC Curve (AUC) with respect to $Y^c$ in different simulations of experts' decisions, corresponding to different assumptions of human expertise and errors. The proposed methodology outperforms all others across the different scenarios.}
\label{tab:sim_auc_err_app}
\end{table}

\newpage

\subsection{Ablation studies}
\label{app:ablation}

In order to assess the value of the different components of the methodology, we conduct ablation studies. 

\subsubsection{Alternative to influence-based estimation}
\label{app:ablation_influence}

First, we assess the value of the influence-based estimation of expert consistency. To do this, we implement an alternative in which we estimate consistency by training separate models for each expert, and then use the predictions across all models to directly estimate whether there is $\delta$-level consistency. Specifically, let ${f}_h$ be a model trained to predict the decisions of expert $h$ for $h \in \{1,...,q\}$, the estimated consistency set is defined as,
\begin{eqnarray}
   \mathcal{A}' =  \{\bm{x} \in \mathbf{X}  : \frac{1}{q}\sum_{h \in \{1,...,q \} } ( {f}_{h}(\bm{x}) \geq 0.5  ) \geq \delta \}
\end{eqnarray}

In our ablation study, we use $\mathcal{A}'$ as an alternative to $\mathcal{A}$. The results are presented in Tables \ref{tab:abl:sim_auc_err},
\ref{tab:abl:sim_auc_bias} and
\ref{tab:abl:sim_tnr_bias}. The results indicate that this variant always underperforms when compared to $f_{\mathcal{A}}$. Furthermore, while it can sometimes outperform both $f_Y$ and $f_D$, it is not robust to settings where humans have shared incorrect homogeneous beliefs, nor when they have any type of shared biases.These results show that the proposed influence function approach yields a better and much more robust performance.

\subsubsection{Alternative to label amalgamation}
\label{app:ablation_amalgamation}

Second, we assess the value of the label amalgamation by considering an alternative. We consider a hybrid model, which can also be understood as an ensemble model with rejection~\citep{chow1957optimum,chow1970optimum}. This approach trains two separate models: one that predicts the expert decisions, $D$, and one that predicts the observed outcomes, $Y$. At prediction time, 
if $\bm{x} \in \mathcal{A}$, the proposed hybrid approach relies on the model trained to predict the expert decision (i.e., it returns the estimate of what experts would do). If $\bm{x} \not\in \mathcal{A}$,
it relies on the model trained to predict the observed outcome. 

Because the hybrid approach defaults to the prediction of the observed outcome only when an instance lies outside of the estimated $\delta$-level consistency set, we can train the model predicting observed outcomes to specialize on these instances. Thus, let ${f}_{Y_{\neg \mathcal{A}}}(\bm{x})$ be a model that is trained on instances outside of the inferred consistency set (i.e., $\{ \bm{x}\in \mathbf{X} \wedge \bm{x} \notin \mathcal{A}\}$). Formally, the prediction of the hybrid model is given by:
%
\begin{equation}
\begin{array}{rcl}
{f}_{hyb} (\bm{x}) & = & \left\{ \begin{array}{rcl}
{f}_D(\bm{x}) & \mathit{if} & \bm{x} \in \mathcal{A} \\
{f}_{Y_{\neg \mathcal{A}}}(\bm{x})& \mathit{if} & \bm{x} \notin \mathcal{A} \\
\end{array}.\right. 
\end{array}
\label{eq:hyb}
\end{equation}
The primary difference between the hybrid model and the label amalgamation model is that in the
hybrid approach, the two models may be entirely different and may make predictions on different grounds. Meanwhile, in the label amalgamation approach, a single model is learned to predict the amalgamated label $Y^\mathcal{A}$. The results in Tables~\ref{tab:abl:sim_auc_err} and~\ref{tab:abl:sim_auc_bias} indicate that this approach can only outperform $f_D$, but cannot succesfully leverage estimated consistency to outperform $f_Y$, illustrating the value of label amalgamation to boost performance.

\begin{table}
\centering
\footnotesize
\begin{tabular}[b]{lccccc}
\midrule
\textit{\textbf{Scenarios}} & ${f}_Y$    & ${f}_D$  & $f_{hyb}$ & ${f}_\mathcal{A'}$ & ${f}_\mathcal{A}$ \\ \hline
Correct homo. & 0.795 (0.003) &  0.545 (0.024) &  0.795 (0.007) &  0.833 (0.004) &   \textbf{0.856} (0.008)\\
Corr. \& inc. homo.& 0.795 (0.003) &  0.437 (0.007) &  0.802 (0.008) &  0.519 (0.010) &   \textbf{0.860} (0.008) \\
Corr. \& inc. hetero.& 0.795 (0.003) &  0.551 (0.010) &  0.795 (0.006) &  0.835 (0.004) &   \textbf{0.860} (0.007) \\\bottomrule \end{tabular}
\captionof{table}{Alternatives' Area under the ROC Curve (AUC) with respect to $Y^c$ in different simulations of experts' decisions, corresponding to different assumptions of human expertise and errors. The proposed methodology outperforms all alternatives across the different scenarios.}
\label{tab:abl:sim_auc_err}
\end{table}

\begin{table}[t]
\centering
\footnotesize
\begin{tabular}[b]{lccccc}
\midrule
\textit{\textbf{Scenarios}} & ${f}_Y$    & ${f}_D$  & $f_{hyb}$ & ${f}_\mathcal{A'}$ & ${f}_\mathcal{A}$ \\ \hline
Det. bias, part. shared                & 0.794 (0.003) &  0.611 (0.013) &  0.790 (0.005) &  0.774 (0.006) &   \textbf{0.805} (0.008) \\
Homo. bias, fully shared & 0.794 (0.003) &  0.612 (0.023) &  0.788 (0.004) &  0.772 (0.004) &   \textbf{0.806} (0.008) \\
Non-rand. assign.            & 0.794 (0.003) &  0.614 (0.008) &  0.790 (0.006) &  0.772 (0.006) &   \textbf{0.802} (0.008) \\ \hdashline
Det. bias shared                & \textbf{0.794} (0.003) &  0.544 (0.002) &  0.542 (0.004) &  0.567 (0.019) &   0.574 (0.042) \\\bottomrule \end{tabular}
\vspace{0.01in}
\captionof{table}{Alternatives' Area under the ROC Curve (AUC) in different simulations of experts' decisions, corresponding to different assumptions about humans' shared bias. All alternative are robust to biases, if these are not deterministic and shared across all experts.}
\label{tab:abl:sim_auc_bias}
\end{table}

\begin{table}[t]
\centering
\footnotesize
\begin{tabular}[b]{lccccccc}
\midrule
\textit{\textbf{Scenarios}}  & ${f}_Y$    & ${f}_D$  & $f_{hyb}$ & ${f}_\mathcal{A'}$ & ${f}_\mathcal{A}$ \\ \hline
Det. bias, part. shared & \textbf{0.189} (0.002) &  0.120 (0.008) &  0.177 (0.004) &  0.162 (0.003) &   0.180 (0.003) \\
Homo. bias, fully shared & \textbf{0.189} (0.002) &  0.109 (0.021) &  0.176 (0.004) &  0.162 (0.003) &   0.180 (0.004) \\
Non-rand. assign. & \textbf{0.189} (0.002) &  0.121 (0.005) &  0.176 (0.004) &  0.163 (0.004) &   0.179 (0.004) \\\hdashline
Det. bias shared    & \textbf{0.189} (0.002) &  0.078 (0.001) &  0.078 (0.001) &  0.092 (0.011) &   0.094 (0.028) \\\bottomrule \end{tabular}
\vspace{0.01in}
\captionof{table}{Alternatives' True Negative Rate (TNR) for women, the group discriminated against in the simulations of experts' decisions, corresponding to different assumptions about humans' shared bias. All alternatives present a TNR similar to $f_Y$ even when $f_D$ presents significantly lower performance.}
\label{tab:abl:sim_tnr_bias}
\end{table}

\subsection{Application of diagnosis of bias risk}
\label{app:bias_risk_sim}

When we have access to an oracle, $Y^c$ in simulations, the ability to assess bias through error rates, as done in Table~\ref{tab:sim_tnr_bias}, is ideal. However, because this assessment is not possible in real-world settings, we also consider the demographic parity assessment $\Delta  \textrm{Gap}_{dp} $ proposed in Equation~\eqref{eq:dem_gap_diff}. This statistic, reported in Table~\ref{tab:sim_dem_bias}, shows the change in selection rates across genders when using ${f}_{\mathcal{A}}$ vs. ${f}_Y$. These results illustrate two things that we discussed conceptually in Section~\ref{app:subsec_bias}. First, we see that a non-zero value is not necessarily indicative of increased error rates for one group, emphasizing the fact that demographic parity gaps suggest a \emph{potential} issue and should not be treated as conclusive evidence. Second, the very large value seen in the failure mode indicates that this diagnostic can serve as a red flag, thus warning about the potential risk of estimated consistency carrying bias. As such, an interpretation of this diagnostic, grounded on domain knowledge, can support the decision of whether learning from estimated expert consistency in a given setting may be desirable.

\begin{table}[h]
\centering
\footnotesize
\begin{tabular}[b]{lc}
\midrule
\textit{\textbf{Scenarios}}   & $\Delta  \textrm{Gap}_{dp} $  \\  \hline
Deterministic bias, partially shared & -0.213 \\
Homogenous bias, fully shared & -0.206 \\
Non-rand. assign. \& near deter. bias &-0.208 \\\hdashline
Deterministic bias shared (fail mode) &  -0.590 \\\bottomrule
\end{tabular}
\vspace{0.01in}
\captionof{table}{$\Delta  \textrm{Gap}_{dp} $ as defined in Equation~\eqref{eq:dem_gap_diff}, across different simulations of humans' shared bias. The proposed diagnostic cases serves to identify the scenario where the method should not be used (fail mode).}
\label{tab:sim_dem_bias}
\end{table}

\newpage
\subsection{Further analysis of non-random expert-to-patient assignment, near-deterministic bias}
\label{app:mimic_hyperparam} 
Among simulation scenarios, the local influence method is most important for the non-random expert-to-patient assignment, near-deterministic bias scenario. This is because in this setting high-probability predictions of human decisions may be driven by a single expert. Figure~\ref{fig:inf_ind} shows the mean influence per expert, estimated across all cases in each subgroup. Expert 0 is the expert that assesses 95\% of women's cases and always screens them out. In this instance, the local influence method shows that one expert is at odds with the rest, and the prediction of low probability of screen-in is driven by a single expert. 

We investigate how this bias impacts the proposed metrics of consistency. Figure~\ref{fig:metrics} shows the metrics differentiated by group. The center of mass, in Figure~\ref{fig:metrics}a, shows that the agreeing proportion of experts is smaller for female than for male patients. The aligned influence, in Figure~\ref{fig:metrics}b, shows that, while influence for male instances is very aligned, this is not the case for female instances. 

Motivated by Figure~\ref{fig:metrics}b, we study the impact that varying $\gamma_2$ has on amalgamation and the resulting model's performance. Table~\ref{fig:gamma} shows the performance of ${f}_{\mathcal{A}}$ when varying this parameter. In these results it can be seen that using a very small parameter value 
leads to the amalgamation of predictions that are heavily influenced by a single (biased) expert, resulting in a model that exhibits bias against the minority group. However, the model presents robust performance on a large range of the parameter.
These results indicate that in this setting the proposed approach enables a good estimation of the set $\mathcal{A}$ that is not overly sensitive to parameter tuning. Moreover, the results underscore the importance of the proposed influence function-based expert consistency estimation as a way to prevent bias in amalgamation.  

\begin{figure}[ht]
\centering
\includegraphics[width=0.6\linewidth,clip]{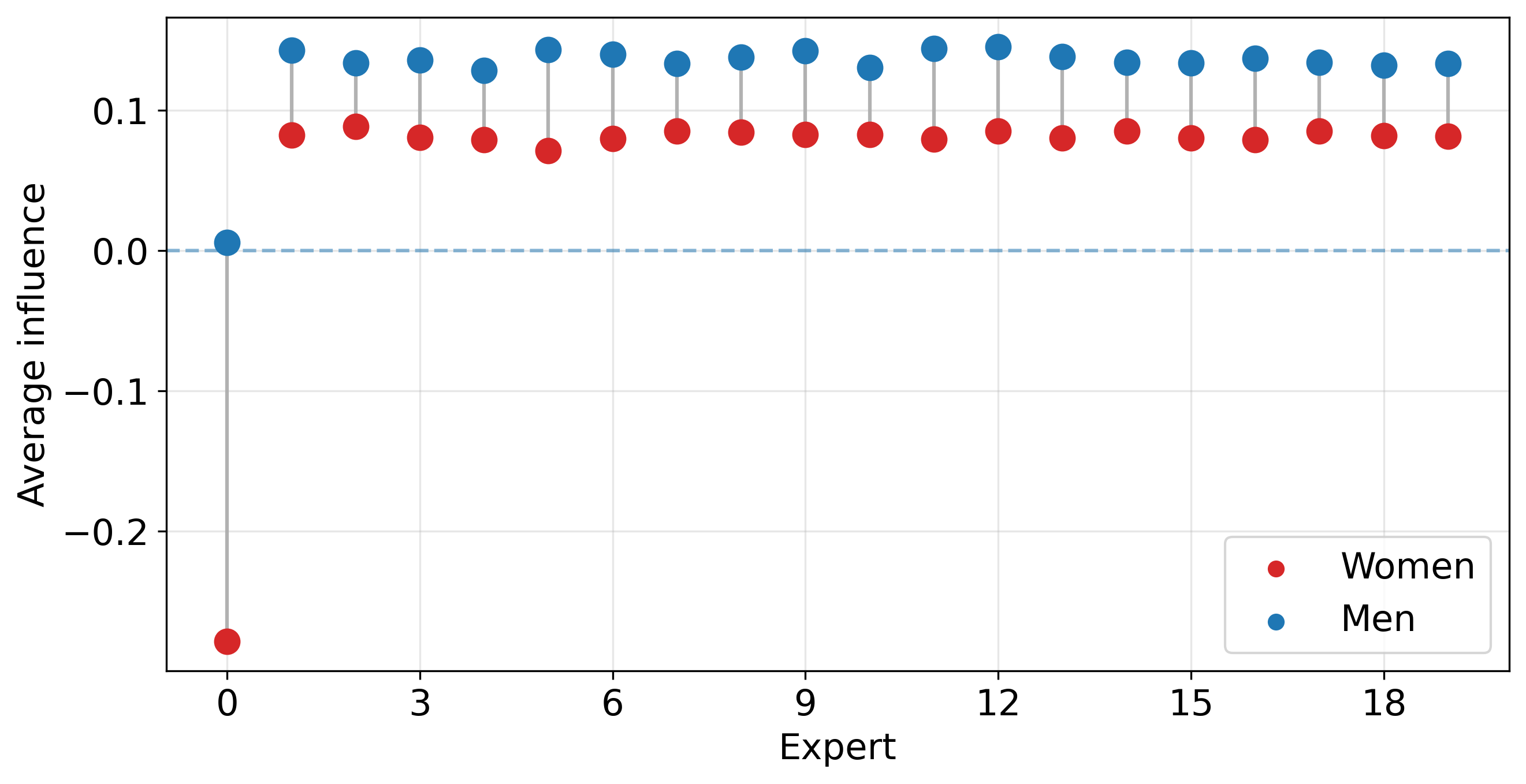}
\caption{Mean influence per expert in non-random expert-to-patient assignment, near-deterministic bias setting, estimated across all cases in each subgroup. Expert 0 is the expert that assesses 95\% of women's cases and always screens them out.}
\label{fig:inf_ind}
\end{figure}

\begin{figure}[ht]
\centering
    \begin{subfigure}{0.4\textwidth}
        \centering
        \includegraphics[width=\textwidth]{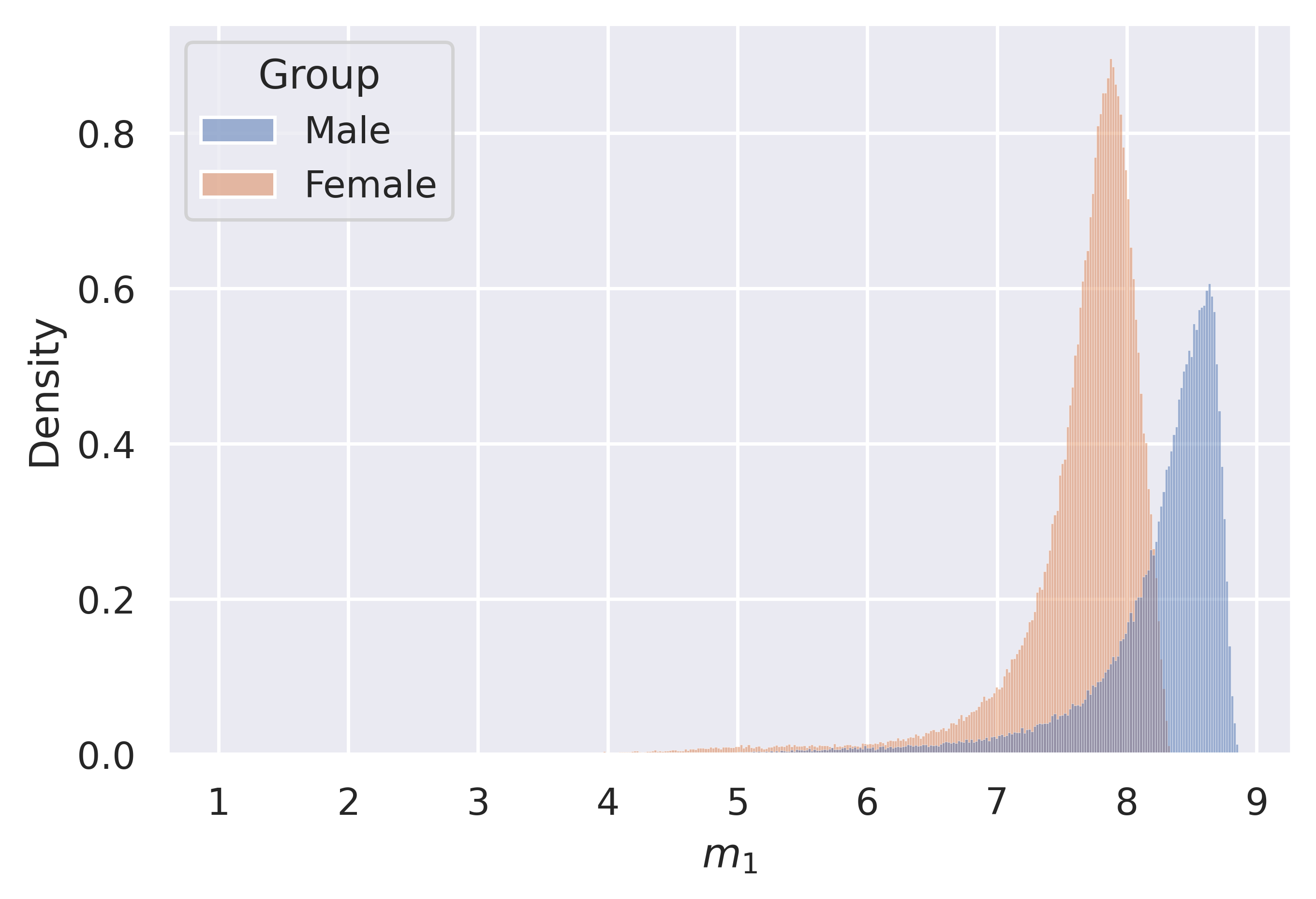}  
        \caption{Center of mass}
        \label{fig:m1}
    \end{subfigure}
    \begin{subfigure}{0.4\textwidth}
        \centering
        \includegraphics[width=\textwidth]{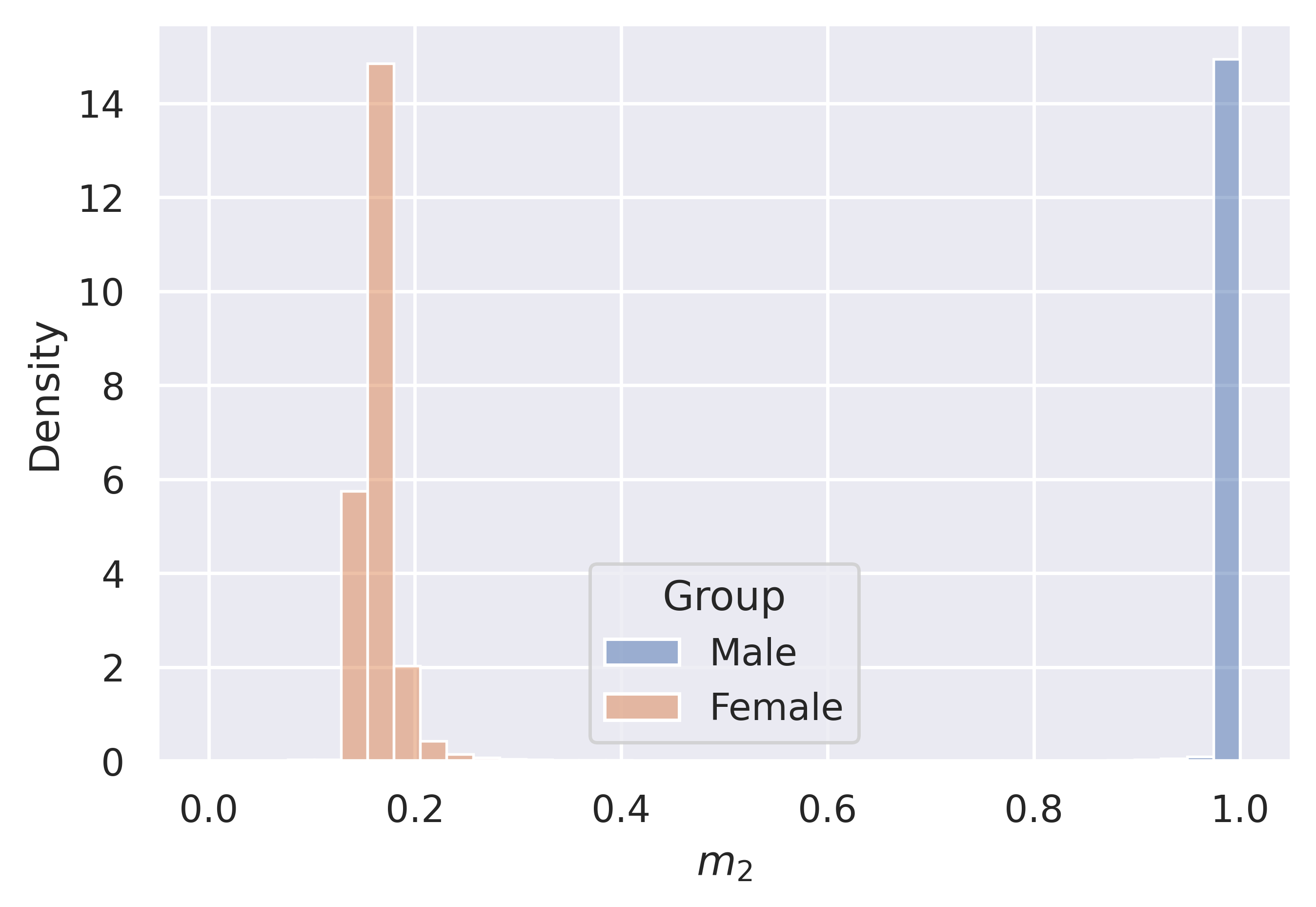}
        \caption{Aligned influence}
        \label{fig:m2}
    \end{subfigure}
    \caption{Measure of consensus in non-random expert-to-patient assignment, near-deterministic bias scenario differentiated by group membership. Our proposed heuristic avoids incorporating biases by relying on multiple complementary metrics.}
    \label{fig:metrics}
\end{figure}

\begin{table}[ht]
\centering
\footnotesize
\begin{tabular}{c|ccccc}
$\gamma_2$& 0.1           & 0.2           & 0.35          & 0.65          & 0.8           \\\hline
AUC & 0.596 (0.010) & 0.799 (0.006) & 0.799 (0.006) & 0.800 (0.005) & 0.799 (0.006) \\
TNR$_{w}$ & 0.117 (0.005) & 0.179 (0.004) & 0.179 (0.004) & 0.180 (0.003) & 0.179 (0.002)
\end{tabular}
\caption{Area Under the ROC Curve (AUC) and True Negative Rate for women (TNR$_{w}$) of ${f}_{\mathcal{A}}$ computed across different values of $\gamma_2$. More inclusive parameters' choice may integrate biases and integrates biases.}
\label{fig:gamma}
\end{table}


\newpage

\subsection{Performance under selective labels problem}
\label{app:mimic_selective}
Below we present analogous experiments to all simulations presented in Section~\ref{subsec:semi}, assuming the presence of the selective labels problem. Here, we assume that outcome $Y$ is only observed when a human decides to screen in a case, as is the case in the real-world example of child maltreatment hotline screenings. We assume that the selective labels problem affects the labels available for training, but still provide evaluation with respect to an oracle, which assesses performance with respect to $Y^c$ for all instances in the test set. Results are presented in Tables~\ref{app:fig:auc_mimic},~\ref{app:fig:auc_mimic_bias} and~\ref{app:fig:tnr_mimic_bias}. First, note that ${f}_D$ and $f_{noise}$ are unchanged, because even under selective labels the human decision is fully observed. All other methods present at least a slight drop in performance. This decrease is particularly pronounced for ${f}_{def}$ and $f_{weak}$, indicating the sensitivity of these methods to the selective labels problem. Their performance is very poor across all settings, and they perform \emph{worse} than both $f_Y$ and $f_D$ when there are shared human biases of any kind. 
$f_{\mathcal{A}}$ performance remains robust, and---consistent with the main results---outperforms all alternatives, except in the case that corresponds to its failure mode. 

\begin{table}
\centering
\footnotesize
\addtolength{\tabcolsep}{-2pt}
\begin{tabular}[b]{lccccccc}
\midrule
\textit{\textbf{Scenarios}} & ${f}_Y$    & ${f}_D$  & ${f}_{def}$  & $f_{noise}$& $f_{ens}$ & $f_{weak}$ & ${f}_\mathcal{A}$ \\ \hline
Correct homo. & 0.782 (0.003) &  0.546 (0.015) &  0.514 (0.011) &  0.778 (0.004) &  0.784 (0.048) &  0.585 (0.004) &   \textbf{0.848} (0.010) \\
Corr. \& inc. homo.&  0.782 (0.003) &  0.435 (0.007) &  0.408 (0.005) &  0.634 (0.010) &  0.775 (0.030) &  0.487 (0.007) &   \textbf{0.846} (0.008) \\
Corr. \& inc. hetero.& 0.781 (0.004) &  0.552 (0.008) &  0.530 (0.008) &  0.795 (0.004) &  0.770 (0.033) &  0.606 (0.003) &   \textbf{0.846} (0.011) \\\bottomrule \end{tabular}
\captionof{table}{Methods' Area under the ROC Curve (AUC) with respect to $Y^c$ in different simulations of experts' decisions, corresponding to different assumptions of human expertise and errors. We assume the selective labels problem constrains labels available for training. The proposed methodology outperforms all others across the different scenarios.}
\label{app:fig:auc_mimic}
\end{table}

\begin{table}[t]
\centering
\footnotesize
\addtolength{\tabcolsep}{-2pt}
\begin{tabular}[b]{lccccccc}
\midrule
\textit{\textbf{Scenarios}}                     &${f}_Y$    & ${f}_D$  & ${f}_{def}$  & $f_{noise}$& $f_{ens}$ & $f_{weak}$ & ${f}_\mathcal{A}$ \\ \hline
Det. bias, part. sh                &  0.772 (0.005) &  0.610 (0.009) &  0.510 (0.050) &  0.652 (0.006) &  0.740 (0.018) &  0.519 (0.011) &   \textbf{0.793} (0.010) \\
Hom. bias, fully sh & 0.778 (0.004) &  0.592 (0.043) &  0.557 (0.029) &  0.678 (0.005) &  0.757 (0.019) &  0.564 (0.008) &   \textbf{0.792} (0.008) \\
Non-rand. assign.            &  0.775 (0.004) &  0.611 (0.010) &  0.527 (0.052) &  0.654 (0.004) &  0.741 (0.016) &  0.521 (0.010) &   \textbf{0.794} (0.014) \\ \hdashline
Det. bias shared                & 0.713 (0.025) &  0.544 (0.002) &  0.506 (0.040) &  0.628 (0.013) &  \textbf{0.740} (0.008) &  0.687 (0.033) &   0.601 (0.022) \\\bottomrule \end{tabular}
\captionof{table}{Methods' Area under the ROC Curve (AUC) in different simulations of experts' decisions, corresponding to different assumptions of human's shared bias.  We assume the selective labels problem constrains labels available for training.}
\label{app:fig:auc_mimic_bias}
\end{table}

\begin{table}[t]
\centering
\footnotesize
\addtolength{\tabcolsep}{-2pt}
\begin{tabular}[b]{lccccccc}
\midrule
\textit{\textbf{Scenarios}}                     &${f}_Y$    & ${f}_D$  & ${f}_{def}$  & $f_{noise}$& $f_{ens}$ & $f_{weak}$ & ${f}_\mathcal{A}$ \\ \hline
Det. bias, part. sh               & 0.170 (0.005) &  0.120 (0.008) &  0.058 (0.034) &  0.140 (0.003) &  0.155 (0.003) &  0.027 (0.005) &   \textbf{0.175} (0.005) \\
Hom. bias, fully sh & 0.173 (0.003) &  0.106 (0.025) &  0.094 (0.020) &  0.146 (0.002) &  0.157 (0.004) &  0.034 (0.002) &   \textbf{0.174} (0.004) \\
Non-rand. assign.            &  0.172 (0.005) &  0.119 (0.007) &  0.070 (0.035) &  0.143 (0.002) &  0.158 (0.004) &  0.027 (0.004) &   \textbf{0.175} (0.003) \\\hdashline
Det. bias shared                & 0.157 (0.011) &  0.078 (0.001) &  0.055 (0.026) &  0.128 (0.004) &  \textbf{0.158} (0.003) &  0.130 (0.028) &   0.112 (0.014) \\\bottomrule \end{tabular}
\captionof{table}{Methods' True Negative Rate (TNR) for women in different simulations of experts' decisions, corresponding to different assumptions of human's shared bias. We assume the selective labels problem constrains labels available for training.}
\label{app:fig:tnr_mimic_bias}
\end{table}

\newpage
\section{Child maltreatment hotline}



\subsection{Logistic regression results}
\label{app:child_log}

Figure~\ref{fig:child_topp_log} shows precision curves when all models rely on logistic regression. Figure~\ref{fig:log_child_25} shows precision at a 25\% cut-off for logistic regression. 
These results echo the conclusions made when using the neural network variant of the models shown in Figure~\ref{fig:child}. Additionally, note how similar the performances of using logistic regression instead of neural netowkrs are for all models, with the exception of ${f}_{def}$, which benefits from a more complex model to estimate the predictions' uncertainty.

\begin{figure}[ht]
\centering
\includegraphics[width=\linewidth]{Management-Science-template/fig/Welfare/log_child.png}
\caption{Precision on child maltreatment risk assessment for top $p\%$ highest scored screened-in cases by model. All models are logistic regression. Error bars show mean$\pm$std over 10 runs of $75-25\%$ Monte Carlo cross-validation. Results show that (1) there are elements of risk that are not captured by the model trained to predict out-of-home placement label ${f}_Y$, but are optimized for by humans, and (2) label amalgamation improves recall and precision for these cases, while having a better performance on out-of-home-placement than a model trained on human decisions alone. }
\label{fig:child_topp_log}
\end{figure}

\begin{figure}[ht!]
\centering
\includegraphics[width=0.8\linewidth,clip]{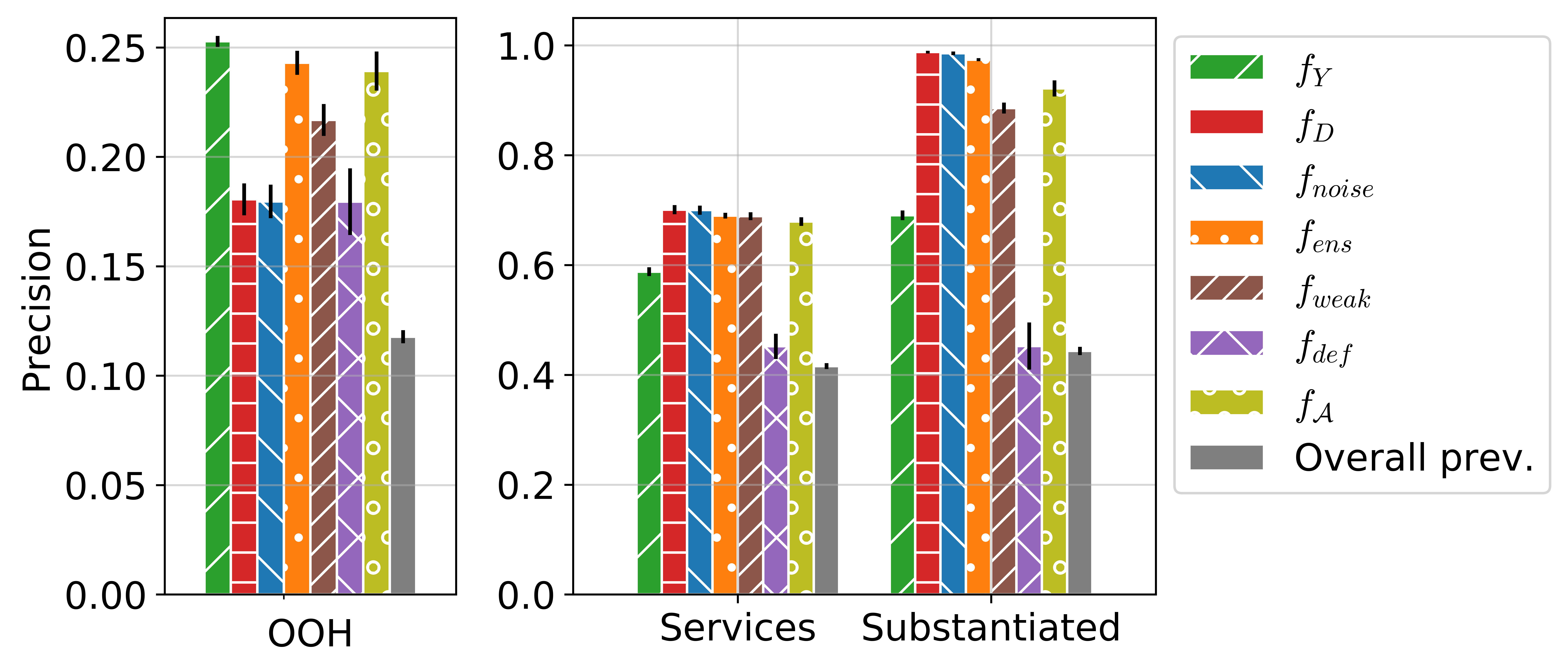}
\caption{Precision on child maltreatment risk assessment for top $25\%$ highest scored screened-in cases by model. }
\label{fig:log_child_25}
\end{figure}

\subsection{Impact of parameters on amalgamation}
\label{app:child_param}
All the presented models on the child maltreatment dataset used parameters $\delta = 0.05$, $\gamma_1 = 4$, $\gamma_2 = 0.8$ and $\gamma_3 = 0.002$. In this section, we provide a sensitivity analysis to these parameters by exploring the impact on the percentage of points amalgamated and on the performance. Figure~\ref{fig:child_hyper} shows the impact on the number of amalgamated points when independently varying each of the parameters, assuming all other parameters remain fixed. These results show that the proposed choice of parameters is conservative. In particular, in our experiments $\gamma_2 = 0.8$, which enforces that at least $80\%$ of the influence is aligned with the dominant direction. Relaxing this parameter could significantly impact the number of amalgamated points.
\begin{figure}[ht]
\centering
    \begin{subfigure}{.49\textwidth}
        \centering
        \includegraphics[width=0.95\textwidth]{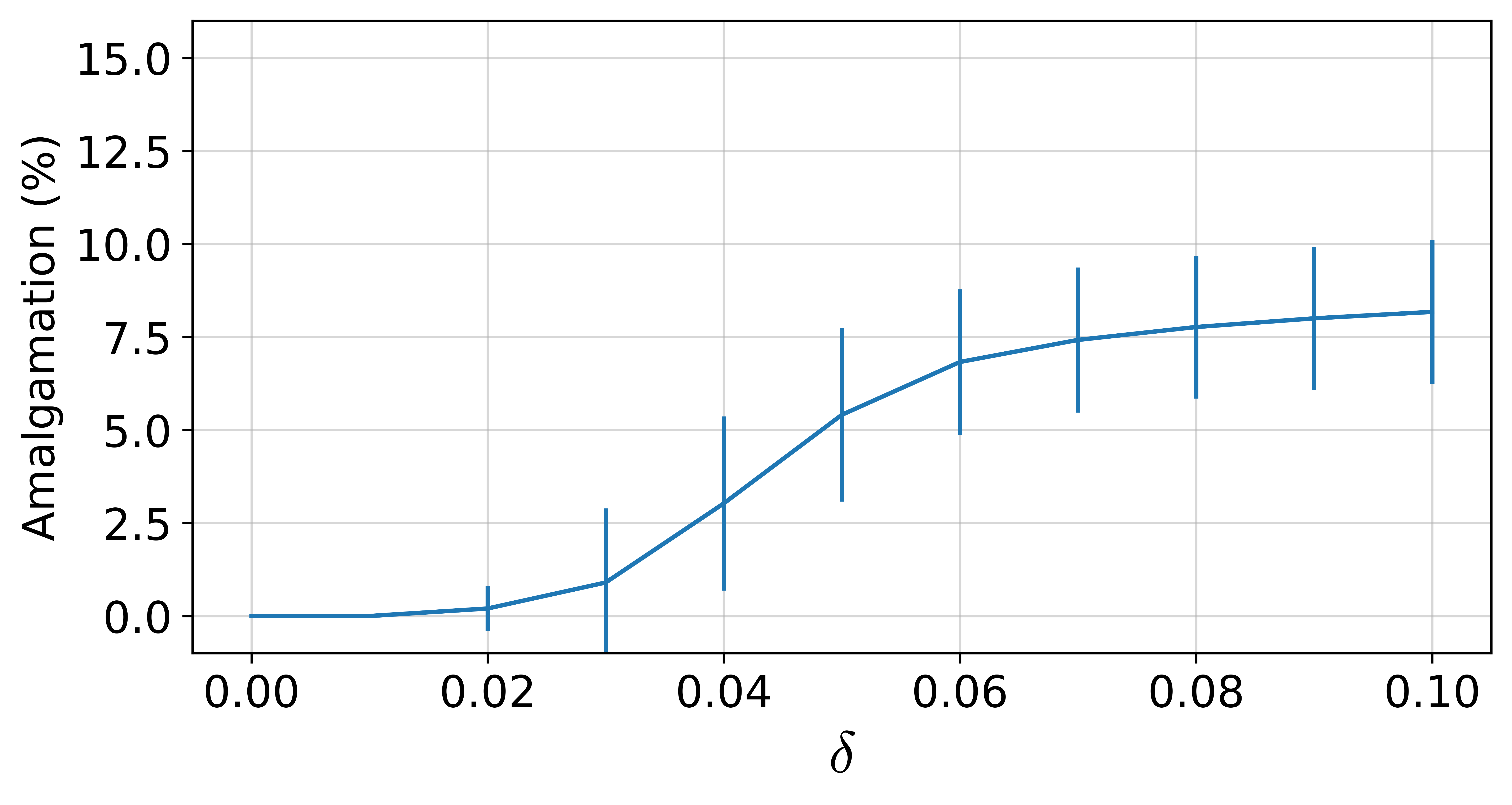}  
        \caption{Amalgamation under different $\delta$}
    \end{subfigure}
    \begin{subfigure}{.49\textwidth}
        \centering
        \includegraphics[width=0.95\textwidth]{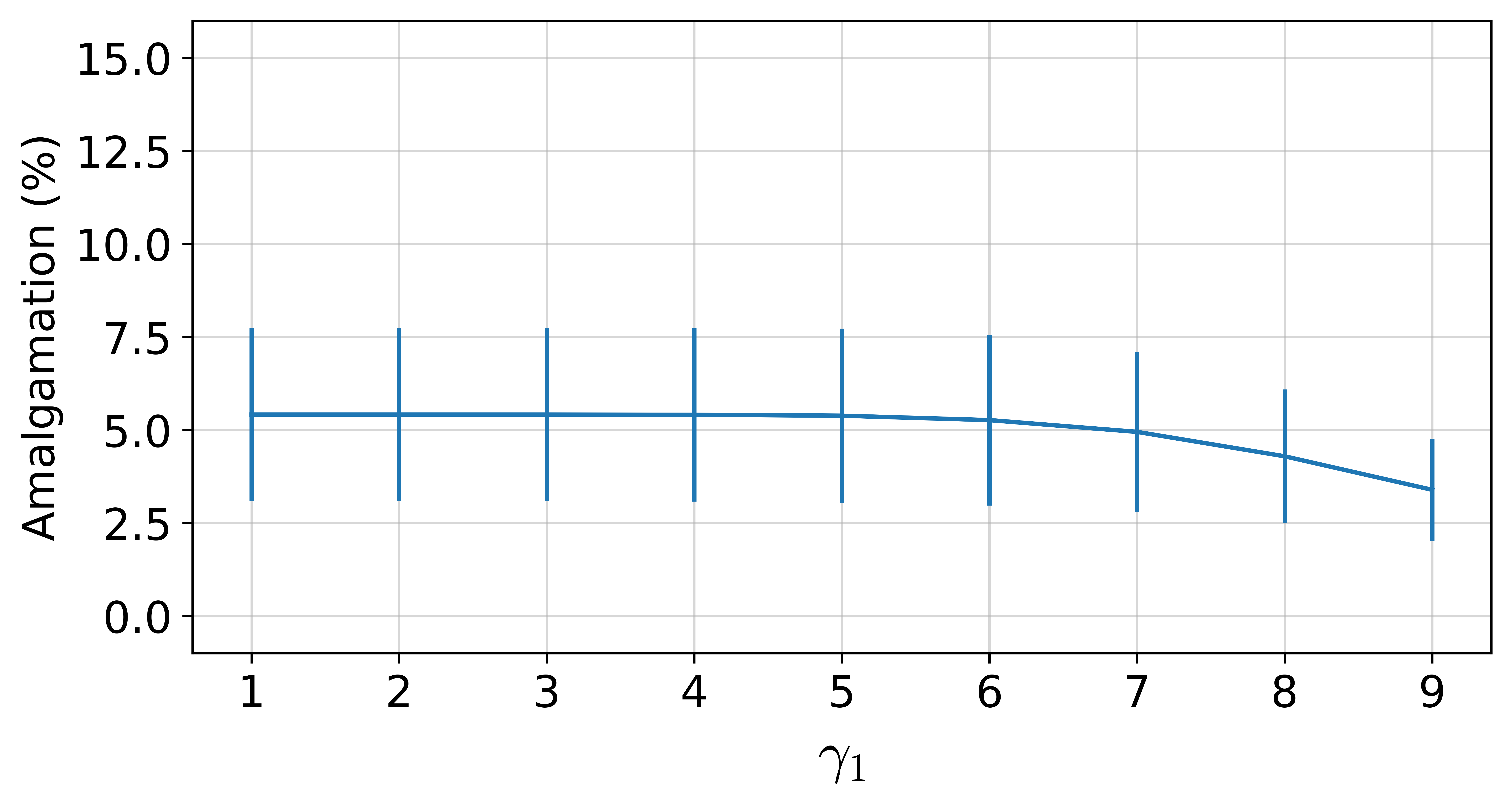}
        \caption{Amalgamation under different $\gamma_1$}
    \end{subfigure}
    \begin{subfigure}{.49\textwidth}
        \centering
        \includegraphics[width=0.95\textwidth]{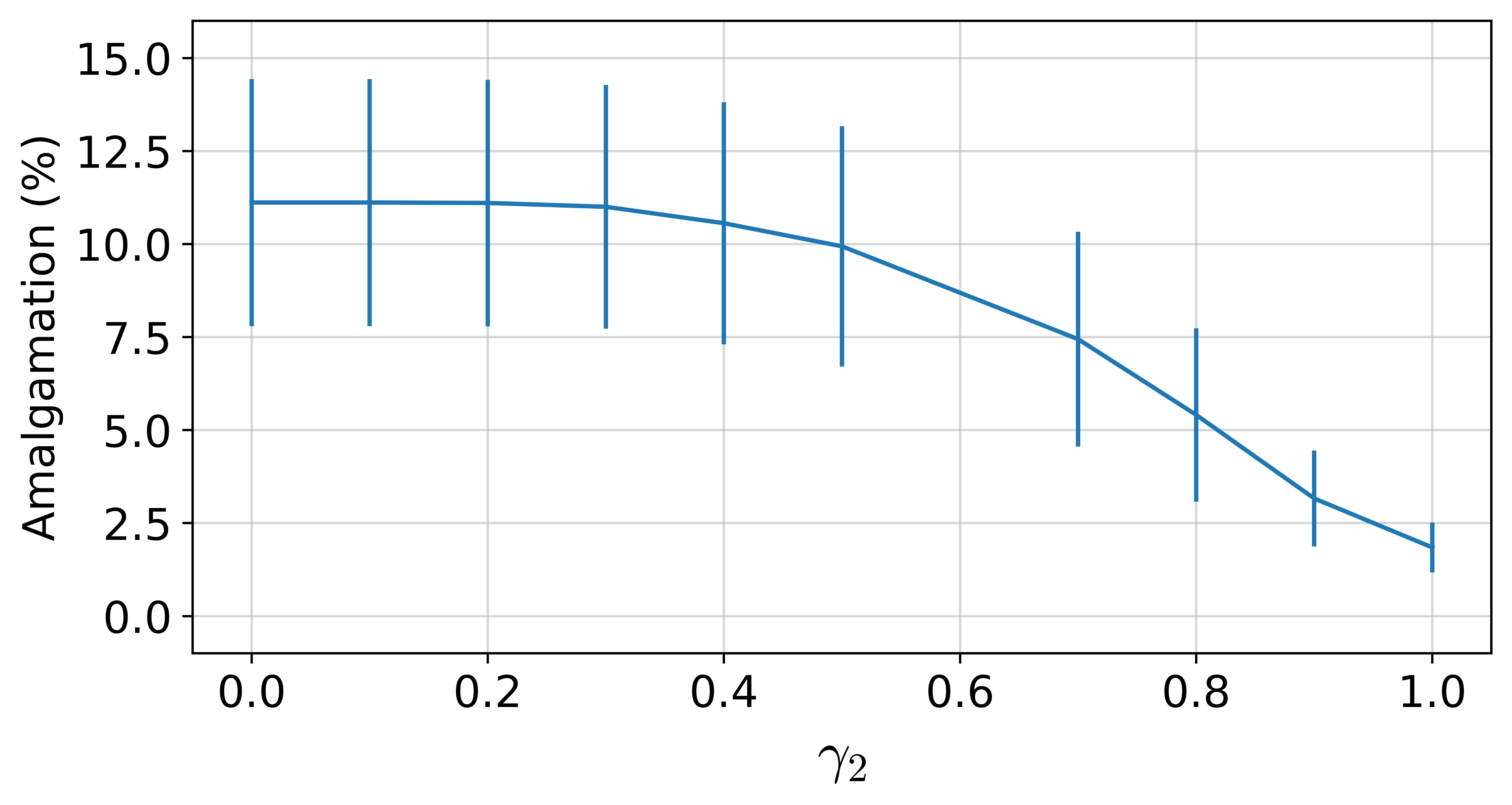}  
        \caption{Amalgamation under different $\gamma_2$}
    \end{subfigure}
    \begin{subfigure}{.49\textwidth}
        \centering
        \includegraphics[width=0.95\textwidth]{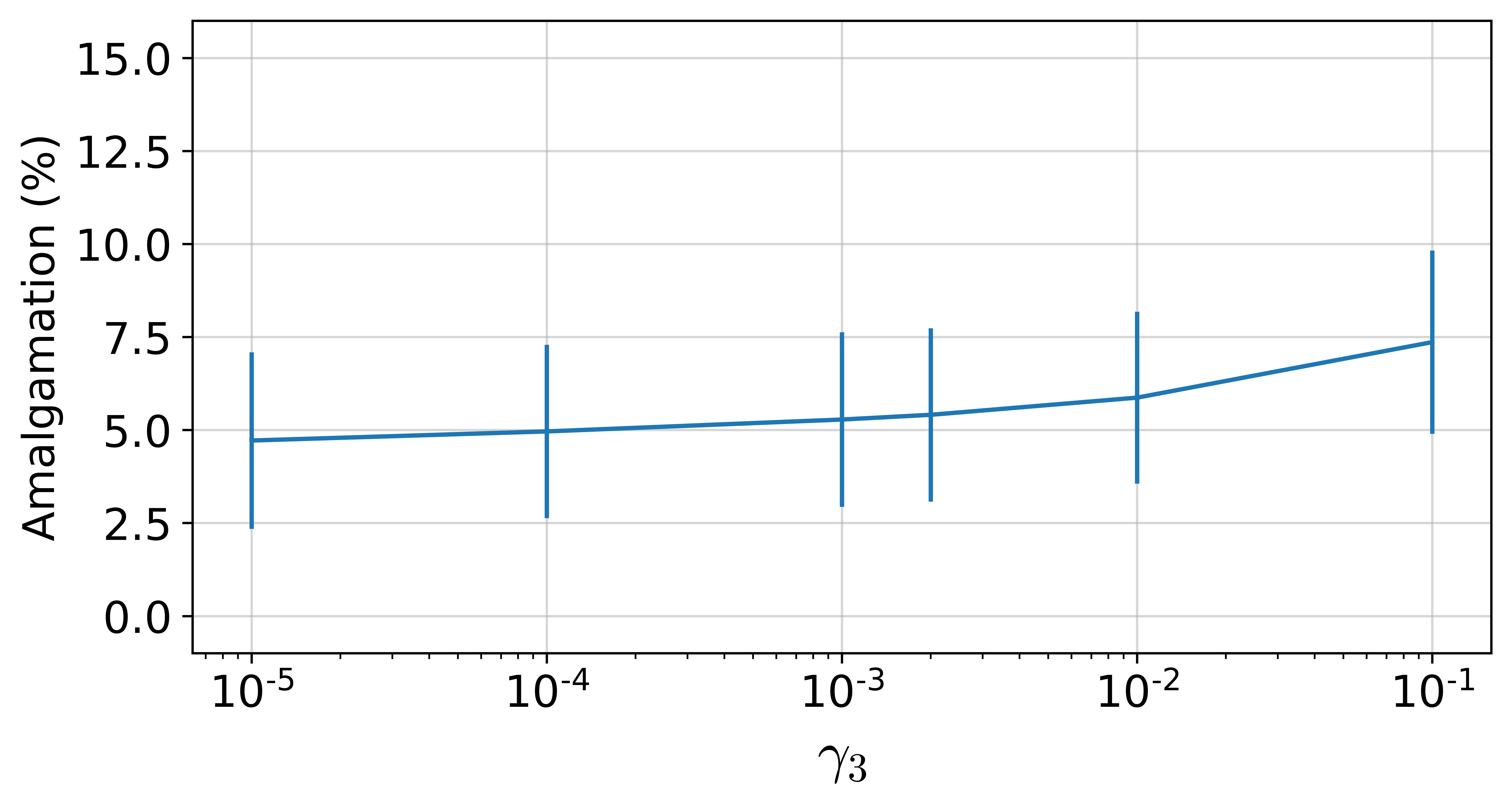} 
        \caption{Amalgamation under different $\gamma_3$}
    \end{subfigure}
    \caption{Impact of parameters on proportion of cases included in the amalgamation set.}
    \label{fig:child_hyper}
\end{figure}

Naturally, correlations between the metrics constrained by different parameters means that varying one parameter while maintaining others at conservative levels may not have a significant effect. In particular, the construction of the set $\mathcal{A}$ relies on the intersection of the two metrics controlled by $\gamma_1$ and $\gamma_2$. When $\gamma_2$ is set to a conservative threshold, this gives the impression that $\gamma_1$ does not impact the amalgamated set. Figures \ref{fig:inde_gamma1} and \ref{fig:inde_gamma2} show the individual impact of $\gamma_1$ and $\gamma_2$ when the other parameter does not constraint the set of amalgamated points, i.e. when $\gamma_2=0$ and $\gamma_1=0$, respectively. This shows that, while both metrics may constrain the set $\mathcal{A}$, in our experiments the amalgamation set is more influenced by $\gamma_2$. This can be explained by the fact that there do not seem to be points for which the influence is dominated by less than four experts.

\begin{figure}[ht]
\centering
    \begin{subfigure}{.49\textwidth}
        \centering
        \includegraphics[width=0.95\textwidth]{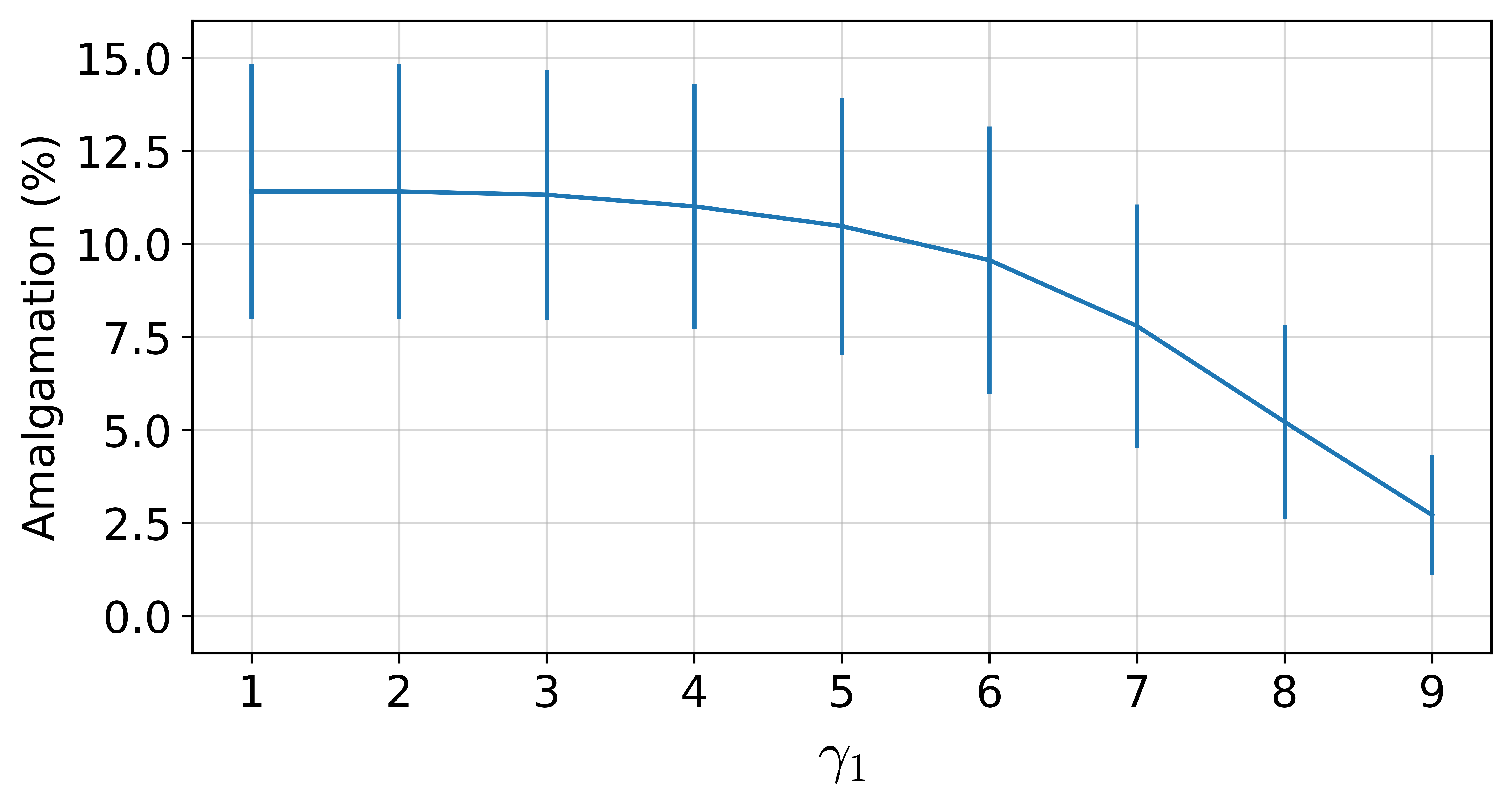}  
        \caption{Individual impact of $\gamma_1$}
        \label{fig:inde_gamma1}
    \end{subfigure}
    \begin{subfigure}{.49\textwidth}
        \centering
        \includegraphics[width=0.95\textwidth]{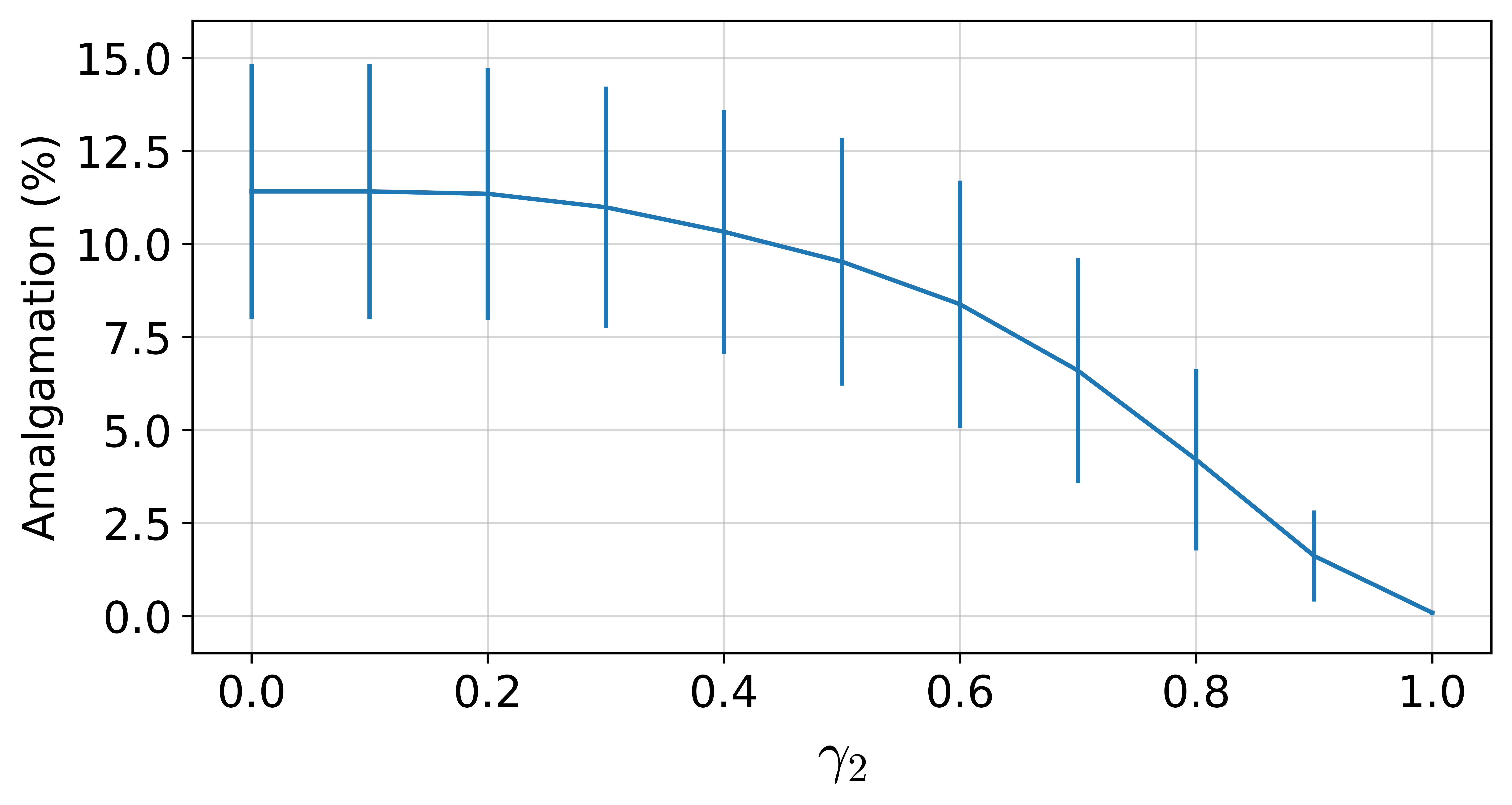}
        \caption{Individual impact of $\gamma_2$}
        \label{fig:inde_gamma2}
    \end{subfigure}
    \caption{Individual impact of parameters on proportion of cases included in the amalgamation set when $\gamma_2 = 0$ in (a) (resp. $\gamma_1 = 0$ in (b)).}
\end{figure}

Finally, we show the method's performance when choosing more extreme sets of parameters, illustrating performance with overly relaxed constraints and overly conservative ones. Figures~\ref{fig:inc} and \ref{fig:cons} shows the precision curves for the quadruplets $(\delta, \gamma_1, \gamma_2, \gamma_3) = (0.05, 2, 0.7, 0.002)$ ---more relaxed--- and $(0.05, 6, 0.9, 0.002)$ ---more conservative. This shows how amalgamating more human decisions results in a model less discriminative for $Y$, but improves the prediction of complementary outcomes. Conversely, relying less on humans leads to better discrimination on $Y$ but hurts performance on the two other outcomes. Across all settings, ${f}_{\mathcal{A}}$ performs significantly better than ${f}_D$ with respect to out-of-home placement, and significantly better that ${f}_Y$ with respect to services and substantiation.

\begin{figure}[ht]
\centering
    \begin{subfigure}{\textwidth}
        \centering
        \includegraphics[width=\textwidth]{Management-Science-template/fig/Welfare/child_evol_inc.png}  
        \caption{With \textit{relaxed} parameters $\gamma_1, \gamma_2$. $(\delta, \gamma_1, \gamma_2, \gamma_3) = (0.05, 2, 0.7, 0.002)$.}
        \label{fig:inc}
    \end{subfigure}
    \begin{subfigure}{\textwidth}
        \centering
        \includegraphics[width=\textwidth]{Management-Science-template/fig/Welfare/child_evol_cons.png}
        \caption{With \textit{conservative} parameters $\gamma_1, \gamma_2$. $(\delta, \gamma_1, \gamma_2, \gamma_3) = (0.05, 6, 0.9, 0.002)$.}
        \label{fig:cons}
    \end{subfigure}
    \caption{Precision on child maltreatment risk assessment for top p\% highest scored screened-in cases by neural network model.}
\end{figure}
\end{APPENDICES}

\end{document}